%% file: main.tex
\documentclass[10pt,twocolumn,letterpaper]{article}
\usepackage{iccv}
\usepackage{multirow}
\usepackage{url}
\usepackage[accsupp]{axessibility}
\definecolor{iccvblue}{rgb}{0.21,0.49,0.74}
\usepackage[pagebackref,breaklinks,colorlinks,allcolors=iccvblue]{hyperref}

\title{SuperMat: Physically Consistent PBR Material Estimation at Interactive Rates}

\author{
Yijia Hong\textsuperscript{1,2}, Yuan-Chen Guo\textsuperscript{4}, Ran Yi\textsuperscript{1}, Yulong Chen\textsuperscript{3}, Yan-Pei Cao\textsuperscript{4}, Lizhuang Ma\textsuperscript{1}\\
$^1$Shanghai Jiao Tong University ~~
$^2$Shanghai Innovation Institute ~~
$^3$Harbin Institute of Technology \\
$^4$VAST
}

\begin{document}
\twocolumn[{%
    \renewcommand\twocolumn[1][]{#1}%
    \maketitle
    \vspace{-0.8cm}
    \begin{center}
        \includegraphics[width=1\linewidth]{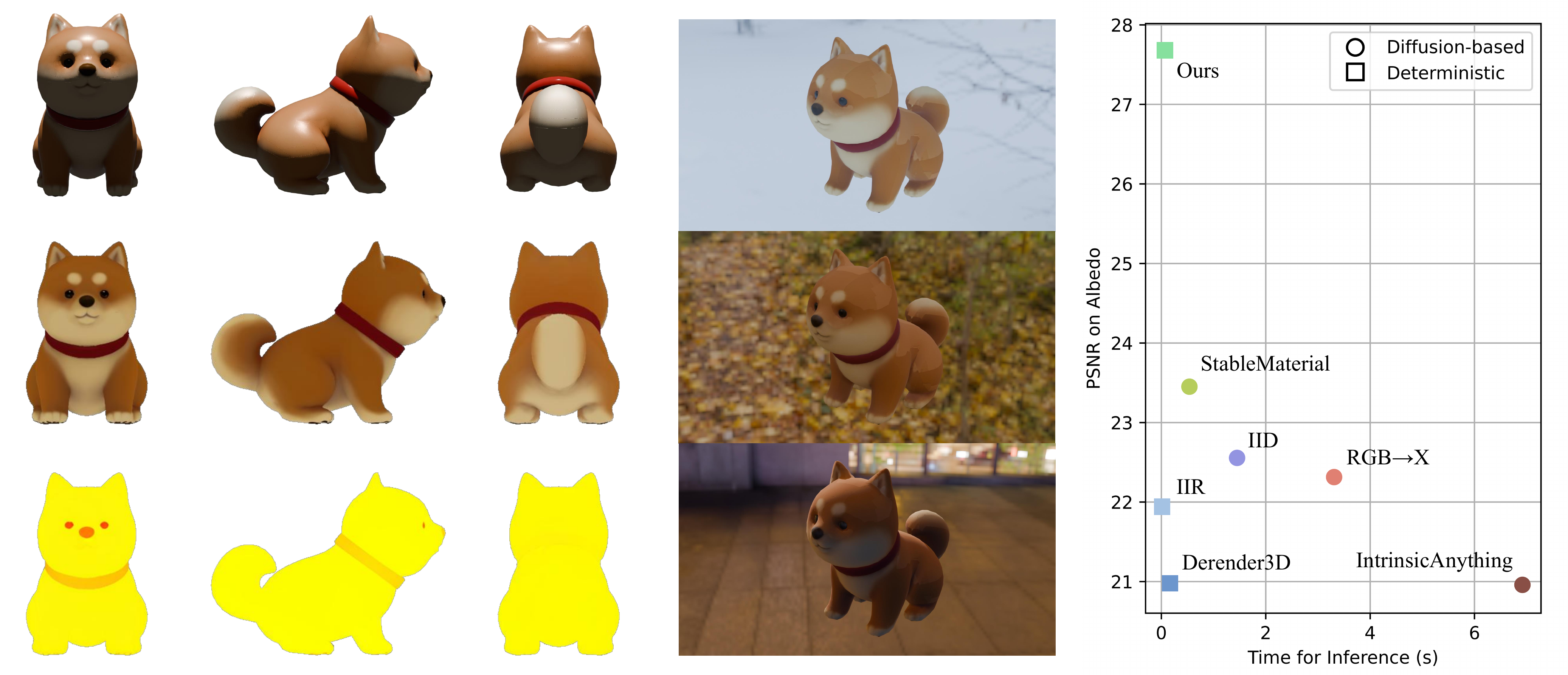}
        \captionof{figure}{We present an efficient method for high-quality material decomposition in both 2D and 3D. \textbf{Left}: Albedo and RM(Roughness, Metallic) predictions of SuperMat on rendered images under unknown illumination. The first row shows inputs with uneven lighting, the second row displays albedo predictions, and the third row presents RM predictions. Each texel in the RM map here is represented as $(1, \text{roughness}, \text{metallic})$. \textbf{Middle}: Rendering results of a 3D object with PBR materials obtained by our method under novel lighting. \textbf{Right}: Our method is efficient and produces high-quality results.}
        \label{fig:7}
    \end{center}
}]

\input{sec/0_abstract}
\input{sec/1_intro}
\input{sec/2_related}
\input{sec/3_method}
\input{sec/4_exper}
\input{sec/5_conc}

{
    \small
    \bibliographystyle{ieeenat_fullname}
    \bibliography{main}
}
\input{sec/X_suppl}

\end{document}

%% file: sec/0_abstract.tex
\begin{abstract}
Decomposing physically-based materials from images into their constituent properties remains challenging, particularly when maintaining both computational efficiency and physical consistency. While recent diffusion-based approaches have shown promise, they face substantial computational overhead due to multiple denoising steps and separate models for different material properties. We present SuperMat, a single-step framework that achieves high-quality material decomposition with one-step inference. This enables end-to-end training with perceptual and re-render losses while decomposing albedo, metallic, and roughness maps at millisecond-scale speeds. We further extend our framework to 3D objects through a UV refinement network, enabling consistent material estimation across viewpoints while maintaining efficiency. Experiments demonstrate that SuperMat achieves state-of-the-art PBR material decomposition quality while reducing inference time from seconds to milliseconds per image, and completes PBR material estimation for 3D objects in approximately 3 seconds. The project page is at \url{https://hyj542682306.github.io/SuperMat/}.
\end{abstract}

%% file: sec/1_intro.tex
\section{Introduction}

Decomposing Physically-Based Rendering (PBR) materials from images into their constituent properties (\textit{e.g.}, albedo, metallic, and roughness maps) remains a central challenge in computer vision and graphics. This task is essential for creating high-fidelity 3D assets across numerous applications, from film production and game development to virtual and augmented reality pipelines. While recent advances have shown promising results \cite{chen2024intrinsicanything, zeng2024rgb, litman2024materialfusion}, the problem remains challenging due to its ill-posed nature, particularly given unknown lighting conditions and complex interactions between different material properties. Our work addresses two key objectives: \textbf{1)} efficient decomposition of high-fidelity material maps from 2D images, and \textbf{2)} extension of this capability to 3D objects, enabling accurate material estimation across complex geometries.

Previous works in PBR material estimation generally fall into two main categories: optimization-based generation for 3D objects and data-driven decomposition for 2D images. Optimization-based methods~\cite{guo2022nerfren, jin2023tensoir, gao2023relightable, liang2024gs} employ differentiable rendering to reconstruct 3D objects with material parameters by minimizing the difference between rendered and target images. Instead, data-driven approaches learn the distribution of various materials from large datasets to perform intrinsic decomposition on 2D images, which is the primary focus of our work. Early methods ~\cite{wimbauer2022rendering, zhu2022learning} in this domain fail to produce stable and reliable results due to limited data availability and model capacity. Recent works ~\cite{kocsis2024intrinsic, zeng2024rgb, chen2024intrinsicanything, litman2024materialfusion} leverage large diffusion models and reframe material decomposition as a conditional image generation task. However, these methods tend to focus solely on individual material properties, overlooking the interactions between different types of materials and the combined impact on rendered appearances. Even minor inaccuracies in individual material estimates can propagate through the rendering pipeline, leading to substantial visual artifacts. 
Moreover, these diffusion-based methods face considerable computational overhead - they demand multiple denoising steps (30-50) during inference, causing a critical bottleneck for real-time applications.

To address these challenges, we propose a deterministic one-step framework that achieves both efficient and high-quality material decomposition. First, we introduce and refine structural expert branches within the UNet architecture. These specialized branches enable concurrent processing of multiple material properties while maintaining domain-specific expertise, allowing our model to decompose albedo, metallic, and roughness maps simultaneously with minimal computational overhead. This design facilitates collaborative and physically consistent material prediction while avoiding the linear cost scaling typically required when estimating multiple material attributes. Second, we address the computational inefficiency of traditional diffusion models by reformulating the DDIM sampling process. Through modification of the scheduler, we achieve single-step inference without compromising quality, dramatically reducing inference time. Most importantly, this single-step formulation enables end-to-end training of the diffusion model, allowing us to compute the perceptual and re-render loss on final predicted material textures, which enhances shadow removal and improves the joint performance of decomposed materials under varying lighting conditions. These advances lead to \emph{SuperMat}, a highly efficient image space material decomposition model that estimates high-fidelity albedo, metallic, and roughness maps \emph{at millisecond-scale speeds}.

For practical 3D applications, we further extend SuperMat's capabilities to handle 3D objects. We upgrade SuperMat to a multi-view model, SuperMatMV, and design a UV refinement network that processes SuperMatMV’s predictions in UV space. This approach addresses the challenge of maintaining consistency and completeness across different, limited viewpoints while preserving the efficiency of our original 2D approach. Our pipeline processes a textured 3D model in approximately \emph{3 seconds}, producing high-quality material maps that maintain consistency across arbitrary viewpoints and lighting conditions.

In summary, our contributions are: 

\begin{itemize}
    \item We propose SuperMat, a novel image space material decomposition model that generates physically consistent albedo, metallic, and roughness maps simultaneously with a single-step inference process. Our method achieves state-of-the-art performance on diverse datasets, significantly outperforming previous diffusion-based works in speed.
    \item With the help of structural expert branches and single-step inference, we develop an end-to-end image space material decomposition training system that generates clean predictions for all material properties in each iteration, allowing for the integration of advanced techniques such as perceptual loss and re-render loss.
    \item We propose an efficient PBR material decomposition method for 3D models. Combining SuperMatMV and our UV refinement network, this pipeline generates high-quality PBR materials for 3D objects from baked textures in approximately 3 seconds.
\end{itemize}

%% file: sec/2_related.tex
\section{Related Works}

\begin{figure*}
    \centering
    \includegraphics[width=1\linewidth]{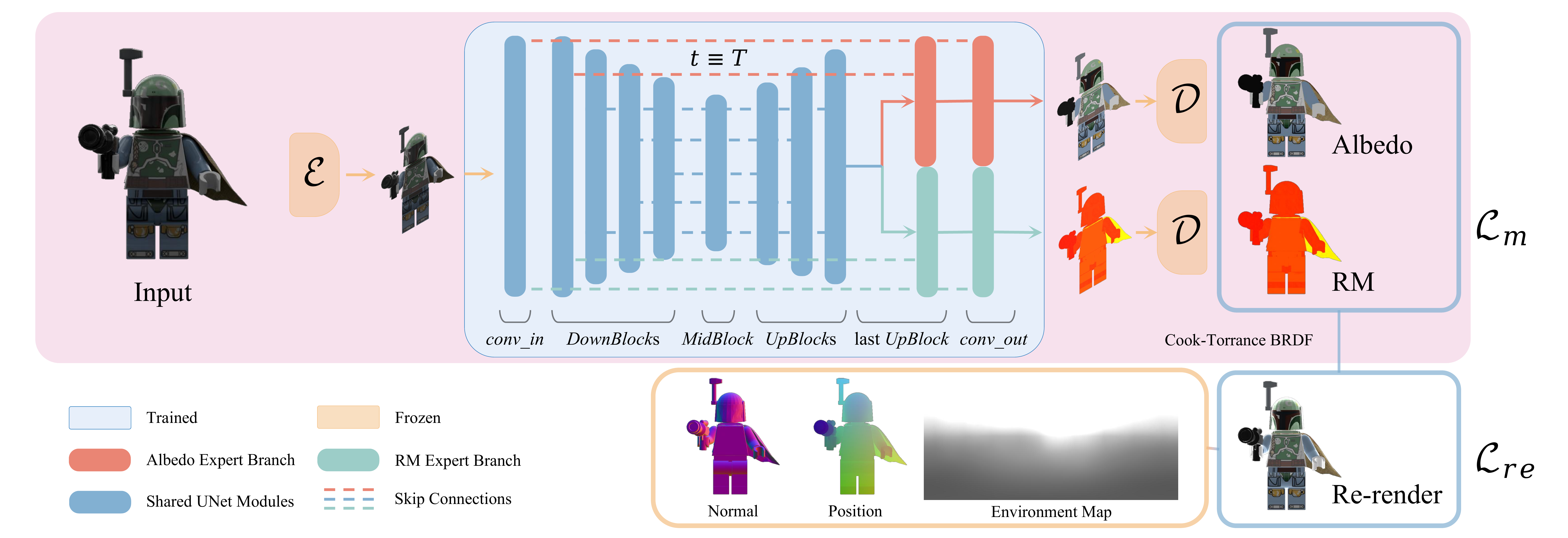}
    \caption{
    Overview of the SuperMat framework. The UNet architecture incorporates structural expert branches for albedo and RM estimation, enabling parallel material property prediction while sharing a common backbone. During training, we leverage a fixed DDIM scheduler and optimize the network end-to-end using both perceptual loss and re-render loss. The \colorbox[HTML]{f7e1ed}{marked} region shows the inference pipeline, where our model achieves single-step inference for concurrent material map generation. Input geometric information (normal, position) and environment maps are used to compute the re-render loss.
    }
    \label{fig:3}
    \vspace{-15pt}
\end{figure*}

\textbf{Material Generation and Decomposition.} Recognizing the importance of material information in 3D assets, researchers have increasingly focused on material acquisition for objects and scenes. On one hand, some 3D generation approaches \cite{chen2023fantasia3d, xu2023matlaber, liu2024unidream, qiu2024richdreamer, xiong2024texgaussian, youwang2024paint, yu2023texture, zhang2024dreammat, deng2024flashtex} integrate material optimization into the appearance creation process, concentrating on generating PBR maps for untextured models. On the other hand, inverse rendering methods aim to reconstruct material-rich 3D representations from single or multi-view images. Some of these methods \cite{zhang2022iron, guo2022nerfren, cheng2023wildlight} assume a predefined lighting model to enhance estimation accuracy, but this constraint limits their practical applicability. To address this, other works \cite{boss2021neural, zhang2021nerfactor, zhang2022modeling, hasselgren2022shape, munkberg2022extracting, jin2023tensoir, gao2023relightable, zhang2024learning, liang2024gs} attempt to recover both material properties and unknown illumination simultaneously. As reliance on predefined assumptions decreases, researchers have found that beyond controlling material decomposition through rendered outputs, directly obtaining image space material priors is also feasible. Building on this idea, similar to ours, another body of research \cite{wimbauer2022rendering, zhu2022learning, kocsis2024intrinsic, zeng2024rgb, chen2024intrinsicanything, litman2024materialfusion, liu2020unsupervised, yi2023weakly, vecchio2024controlmat, wang2024boosting, shi2023attention} focuses on material decomposition from 2D images captured or rendered under unknown lighting conditions. However, existing methods still face challenges related to training data, model architecture, and training strategies, leaving significant room for improvement in decomposition quality.

\noindent\textbf{Diffusion Models.} Recently, researchers have applied diffusion models to a range of tasks \cite{poole2022dreamfusion, chen2023fantasia3d, wang2024prolificdreamer, qiu2024richdreamer, zhang2024dreammat}, demonstrating the effectiveness of casting outcome prediction as a conditional image generation process. By leveraging prior knowledge from large pre-trained models, recent studies \cite{gui2024depthfm, saxena2023monocular, ke2024repurposing, duan2023diffusiondepth, zhao2023unleashing, fu2025geowizard, ye2024stablenormal, chen2024intrinsicanything, litman2024materialfusion} have achieved performance improvements over previous methods in tasks like depth, normal, and material estimation. However, incorporating large models introduces efficiency challenges, particularly high memory requirements and the time-intensive, multi-step denoising process needed for inference. Inspired by insights from recent work \cite{liu2023hyperhuman, garcia2024fine, lin2024common, he2024lotus}, our approach addresses these issues by significantly reducing memory and inference time, improving the feasibility of applying these models in real-world applications.

\noindent\textbf{Generation in UV Space.} UV maps are advantageous for fully encapsulating the appearance information of 3D models, leading several works to explore 3D object manipulation directly within the UV space. By representing the texture of an object as a learnable image, these methods \cite{gao2021tm, mohammad2022clip, richardson2023texture, chen2023text2tex, zeng2024paint3d, boss2024sf3d} enable texture generation, inpainting, and editing, providing new avenues for 3D tasks and allowing 2D image-space techniques and insights to be applied to 3D appearances. Building on this approach, recent methods \cite{bensadoun2024meta, liu2023text, zeng2024paint3d, youwang2024paint, yu2024texgen, yu2023texture, feng2024arm} have applied diffusion models to refine UV maps with promising results. In this spirit, to extend SuperMat's capability to 3D, we fine-tune a UV-space refinement diffusion model specifically.

%% file: sec/3_method.tex
\section{Method}

In this paper, we propose \textbf{SuperMat}, an 
image space material decomposition model fine-tuned from Stable Diffusion that efficiently generates high-quality albedo, metallic, and roughness maps simultaneously.
Given a rendered or captured image of an object under an unknown lighting condition from a specific viewpoint, SuperMat generates the corresponding material maps for the model from that same viewpoint. As shown in Figure \ref{fig:3}, SuperMat is a 2D image-to-image one-step model with \textbf{structural expert branches} that simultaneously generates both albedo and RM maps, supporting \textbf{single-step inference}. Compared to previous diffusion-based methods, this model produces superior results in less time and with reduced memory usage. 
Our single-step formulation allows direct generation of material maps during training, in contrast to prior diffusion methods that predict noise ($\epsilon$) or velocity (v) terms. This enables \textbf{end-to-end} optimization with both perceptual loss on the generated maps and \textbf{re-render loss} from novel lighting conditions, leading to improved material quality and rendering fidelity. Building upon the single-view model, we further introduce a multi-view version, SuperMatMV, to ensure consistency in predictions across different viewpoints.

\begin{figure*}
    \centering
    \includegraphics[width=1\linewidth]{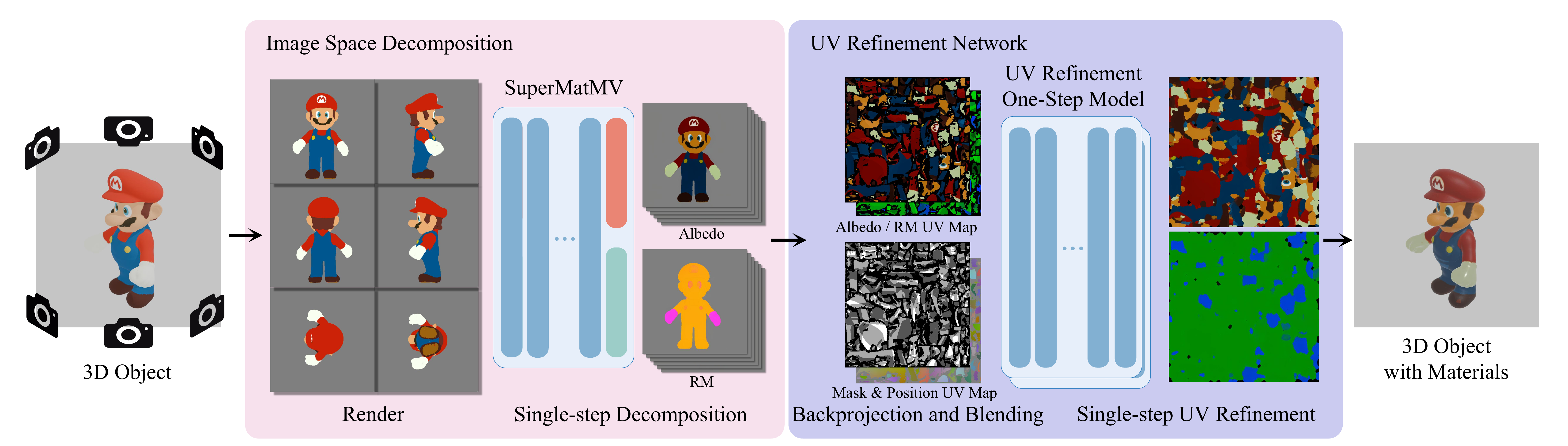}
    \caption{
    Material Decomposition for 3D objects.
    Applying SuperMat's powerful decomposition to 3D objects, we can obtain an object with high-quality materials in just three inference steps. Due to implementation differences, the texels in the RM map for image space decomposition are $(1, \text{roughness}, \text{metallic})$, while in the UV refinement process, they are $(0, \text{roughness}, \text{metallic})$.}
    \label{fig:4}
    \vspace{-15pt}
\end{figure*}

While SuperMat and SuperMatMV achieve efficient image space material decomposition, extending this capability to 3D objects presents additional challenges. Instead of relying on optimization-based approaches like score distillation sampling (SDS) that are computationally intensive and prone to mode collapse, we propose an efficient two-stage pipeline for material estimation on 3D objects. Given a 3D model with only RGB texture, we first apply SuperMatMV to decompose materials from multiple viewpoints. We then introduce a UV refinement network that consolidates these view-dependent predictions into complete, high-quality UV texture maps through single-step inference. As shown in Figure \ref{fig:4}, this pipeline enables PBR material estimation for 3D objects in approximately 3 seconds per model.

\subsection{SuperMat - Image Space Material Decomposition}

Previous works \cite{zeng2024rgb, chen2024intrinsicanything, litman2024materialfusion} have demonstrated that casting material estimation as a conditional latent diffusion process, leveraging knowledge from large pre-trained diffusion models like Stable Diffusion \cite{rombach2022high}, is effective. However, for our task that requires the simultaneous generation of multiple material maps, three key issues remain: 1) High model overhead: Fine-tuning Stable Diffusion with its original architecture can only estimate a single material type. Even when combining metallic and roughness into an RM map, treating it as a single material still requires two diffusion models, one for albedo and another for RM, to meet the rendering model’s requirements. This results in doubled training time and increased inference space overhead. 2) Slow inference speed: With the widely used DDIM scheduler, diffusion models require 30 to 50 denoising steps to produce satisfactory results, which can be inefficient and time-consuming. 3) Insufficient decomposition effect: Due to the high computational cost of backpropagation in multi-step inference, prior diffusion-based methods trained on predicted noise from single-step denoising rather than directly supervising the fully denoised material output. This limitation restricts the model's ability to specialize in the decomposition task and impedes the application of more task-specific techniques.

\noindent\textbf{Structural Expert Branches.} Our solution to the first problem is to introduce a structural expert branch architecture \cite{liu2023hyperhuman, zhang2024clay}. Previous methods are limited to using a single diffusion model to estimate one material type. Although training separate models for different material maps is an intuitive approach, it introduces issues such as complex implementation, high overhead, and inconsistencies between materials. Considering that, despite their different representations, materials fundamentally reflect an understanding of object features, we incorporate structural expert branches into the UNet model, allowing a single diffusion model to predict multiple material maps. Specifically, we design two expert branches, one for predicting albedo, and the other for predicting RM.
For each expert branch, we replicate the UNet's last \textit{UpBlock}, and \textit{conv\_out}, while the rest of the UNet architecture remains shared between the branches. The modified model first extracts and transforms encoded image features through the shared modules. These features are then separated into 2 copies by duplicating before the last \textit{UpBlock}, with each branch ultimately predicting results for its respective domain.
The expert branches specialize in material-specific features, while the shared modules focus on commonalities. In this way, we only add a small number of parameters (19,313,604 parameters, 2.23\% of UNet's total), achieving the same performance as previous methods that would otherwise require twice the scale.

\noindent\textbf{Single-Step Inference and End-to-End Training.} To address the second problem, we analyze the DDIM scheduler of diffusion models. Previous works \cite{garcia2024fine, lin2024common, xu2024diffusion, he2024lotus} have highlighted a flaw in the implementation of the DDIM scheduler, which limits the potential of single-step inference in Stable Diffusion: In the default \texttt{leading} timestep setting of DDIM, for single-step prediction, the model is provided with a timestep ($t\text{=}1$) suggesting the input is essentially denoised, yet in reality, the input is pure noise. The discrepancy between the timestep and the actual noise level causes the single-step result to be flawed. To align the timestep with the noise level received by the model, we adjust the scheduler to use the \texttt{trailing} setting ($t \text{=} T$ for single-step), correcting this issue and enabling single-step inference, significantly reducing the time overhead. 

More importantly, after this correction, we can fine-tune the one-step model end-to-end, effectively resolving the third issue.
 For a diffusion model with the uncorrected timestep setting,
fine-tuning the diffusion model end-to-end is impractical, because generating a clean material map typically requires multiple denoising steps, leading to prohibitive computational complexity due to the gradient calculations. As an alternative, previous methods \cite{zeng2024rgb, chen2024intrinsicanything, litman2024materialfusion} rely on using the noise latent obtained from a single denoising step of the UNet for loss calculation. However, this approach shifts the model's training objective from material decomposition to general image denoising, which hinders specialization for this task. In contrast, our corrected inference pipeline can generate fully denoised material maps in a single denoising step, making end-to-end fine-tuning both feasible and efficient for material decomposition. Specifically, with the clean material estimations, 
we compute perceptual losses \cite{johnson2016perceptual} between the predicted albedo map $k_d$, the RM map $k_{rm}$, and the ground truth maps $\hat{k}_d$, $\hat{k}_{rm}$:
\begin{equation}
\begin{split}
    \mathcal{L}_p(\hat{y}, y) = \sum_{j} \frac{1}{C_jH_jW_j}||\phi_j(\hat{y}) - \phi_j(y)||_1, \\
    \mathcal{L}_m = \mathcal{L}_p(\hat{k}_d, k_d) + \mathcal{L}_p(\hat{k}_{rm}, k_{rm}),
\end{split}
\end{equation}
\noindent
where $\phi$ is a 16-layer VGG network, and $\phi_j$ represents the output of the $j$-th layer, with a shape of $C_j \times H_j \times W_j$. 

Note that, following the approach of previous work \cite{garcia2024fine}, we use “Zeros” as the noise type during fine-tuning, meaning no random noise is added. 

\noindent\textbf{Re-render Loss in DM Fine-tuning.} Data-driven decomposition diffusion models typically focus on the distribution of individual material maps, yet they often overlook the interactions between different types of material maps. In other words, previous methods compute the loss between the predicted values and the ground truth without utilizing material information for further operations or calculations, which could provide stronger guidance for the model. As a result, although the predicted materials may appear close to the ground truth, the rendered output may not align with expectations. This limitation is inherent to their training methods, as they either produce meaningless predicted noise, preventing the application of image-level loss, or fail to generate all material predictions simultaneously. In contrast, our design, which incorporates structural expert branches and single-step prediction, allows us to obtain all denoised material maps during the training iterations. Using this information, we introduce the re-render loss \cite{deschaintre2018single}, which involves rendering the model in a new lighting environment and comparing the result to the ground truth. Specifically, at each training iteration, we obtain $k_d$, $k_{rm} = (k_r, k_m)$, along with the true normal map $\hat{k}_n$, position map $\hat{k}_p$, camera viewpoint $\hat{c}$, and an environment light map. 
Then we use this information and our rendering model $\mathcal{R}$ to compute the predicted material's performance under new illumination. Similarly, with the groundtruth albedo $\hat{k}_d$ and RM $\hat{k}_{rm}$, we compute the ground truth of relighting. The perceptual loss between these two renderings defines our re-render loss:
\begin{equation}
    \mathcal{L}_{re} = \mathcal{L}_p(\mathcal{R}_{\hat{k}_n, \hat{k}_p, \hat{c}}(\hat{k}_d, \hat{k}_m, \hat{k}_r), \mathcal{R}_{\hat{k}_n, \hat{k}_p, \hat{c}}(k_d, k_m, k_r)).
\end{equation}

\noindent\textbf{Extension to Multi-view Material Decomposition. } We extend SuperMat to a multi-view model to enhance the consistency of generated results across different viewpoints. Specifically, we use a 3D self-attention layer and incorporate camera extrinsics as conditions, following MVDream \cite{shi2023mvdream}. The multi-view model, SuperMatMV, is capable of simultaneously decomposing materials for 6 input views, providing more accurate and consistent material predictions.

\subsection{UV Refinement Network}
\label{sec:3.3}

\begin{table*}[ht]
\renewcommand{\tabcolsep}{0.7pt}
\small
\centering
\begin{tabular}{lcccccccccccc|c}
\toprule
                   & \multicolumn{3}{c}{Albedo} & \multicolumn{3}{c}{Metallic} & \multicolumn{3}{c}{Roughness} & \multicolumn{3}{c|}{Relighting} & Time(s)↓ \\ \cline{2-13}
                   & PSNR↑    & SSIM↑  & LPIPS↓ & PSNR↑    & SSIM↑   & LPIPS↓  & PSNR↑     & SSIM↑   & LPIPS↓  & PSNR↑     & SSIM↑    & LPIPS↓   &          \\ \hline
Derender3D \cite{wimbauer2022rendering}        & 20.97 & 0.8606 & 0.1821 & -        & -       & -       & -         & -       & -       & -         & -        & -        & 0.16     \\
IIR \cite{zhu2022learning}                & 21.94 & 0.8709 & 0.1552 & 17.95 & 0.7619 & 0.2911 & 19.73 & 0.8639 & 0.1596 & 20.98  & 0.8884  & 0.1130  & \colorbox[HTML]{ffaeb0}{0.04}     \\ \hline
IID \cite{kocsis2024intrinsic}               & 22.56  & 0.8908  & 0.1282  & 17.38  & 0.7615  & 0.2956  & 20.93  & 0.8759 & 0.1476  & 22.45  & 0.9044  & 0.1083  & 1.45     \\
RGB$\rightarrow$X \cite{zeng2024rgb} & 22.30 & 0.8860 & 0.1154 & 15.36  & 0.7556  & 0.3360  & 20.40  & 0.8473  & 0.1591  & 21.51  & 0.8861  & 0.1132  & 3.32     \\
IntrinsicAnything \cite{chen2024intrinsicanything} & 20.94 & 0.8788 & 0.1520 & -        & -       & -       & -         & -       & -       & -         & -        & -        & 6.93     \\
StableMaterial \cite{litman2024materialfusion}    & 23.44 & 0.9019 & 0.1076 & 20.29 & 0.8164 & 0.2459 & 21.01 & 0.8880 & 0.1327 & 22.56 & 0.9043 & 0.0979  & 0.53     \\
StableMaterialMV \cite{litman2024materialfusion}  & 24.56 & 0.9091 & 0.0946 & 17.28 & 0.7886 & 0.2829 & 20.86 & 0.8838 & \colorbox[HTML]{ffffa8}{0.1286} & 23.14 & 0.9090 & 0.0933 & 0.55 \\ \hline
SuperMat w/o e2e   & 24.26 & 0.8964 & 0.1095 & 20.79 & \colorbox[HTML]{ffffa8}{0.8444} & 0.2826 & 20.81 & 0.8621 & 0.1587 & 23.90 & 0.9085 & 0.0964 & 3.09     \\
SuperMat w/o re-render & \colorbox[HTML]{ffffa8}{26.70} & \colorbox[HTML]{ffffa8}{0.9236} & \colorbox[HTML]{ffffa8}{0.0906} & \colorbox[HTML]{ffffa8}{24.54} & 0.8259 & \colorbox[HTML]{ffffa8}{0.2263} & \colorbox[HTML]{ffffa8}{23.52} & \colorbox[HTML]{ffffa8}{0.9016} & 0.1362 & \colorbox[HTML]{ffffa8}{26.41} & \colorbox[HTML]{ffffa8}{0.9378} & \colorbox[HTML]{ffffa8}{0.0702} & \colorbox[HTML]{ffd7ac}{0.07}     \\
SuperMat           & \colorbox[HTML]{ffaeb0}{27.68} & \colorbox[HTML]{ffaeb0}{0.9290} & \colorbox[HTML]{ffaeb0}{0.0820} & \colorbox[HTML]{ffd7ac}{25.48} & \colorbox[HTML]{ffd7ac}{0.8685} & \colorbox[HTML]{ffd7ac}{0.2221} & \colorbox[HTML]{ffd7ac}{24.25} & \colorbox[HTML]{ffd7ac}{0.9057} & \colorbox[HTML]{ffd7ac}{0.1281} & \colorbox[HTML]{ffaeb0}{27.66} & \colorbox[HTML]{ffaeb0}{0.9432} & \colorbox[HTML]{ffaeb0}{0.0612} & \colorbox[HTML]{ffd7ac}{0.07} \\
SuperMatMV         & \colorbox[HTML]{ffd7ac}{27.56} & \colorbox[HTML]{ffd7ac}{0.9289} & \colorbox[HTML]{ffd7ac}{0.0839} & \colorbox[HTML]{ffaeb0}{26.11} & \colorbox[HTML]{ffaeb0}{0.8749} & \colorbox[HTML]{ffaeb0}{0.2195} & \colorbox[HTML]{ffaeb0}{24.84} & \colorbox[HTML]{ffaeb0}{0.9076} & \colorbox[HTML]{ffaeb0}{0.1208} & \colorbox[HTML]{ffd7ac}{27.64} & \colorbox[HTML]{ffd7ac}{0.9431} & \colorbox[HTML]{ffd7ac}{0.0628} & \colorbox[HTML]{ffffa8}{0.09} \\
\bottomrule
\end{tabular}
\caption{Comparison of our method with others on the image space decomposition task.
We highlight the \colorbox[HTML]{ffaeb0}{best}, \colorbox[HTML]{ffd7ac}{second-best}, and \colorbox[HTML]{ffffa8}{third-best} results for each metric. "SuperMat w/o e2e" denotes our method without the scheduler fix and perceptual loss for end-to-end training. "SuperMat w/o re-render" denotes our method without the re-render loss. The first two methods are non-diffusion-based, while the middle five methods are diffusion-based.}
\label{tab:1}
\vspace{-15pt}
\end{table*}

We aim to apply SuperMat or SuperMatMV to decompose material for 3D objects with only RGB textures.
Several previous works \cite{poole2022dreamfusion, chen2023fantasia3d, zhang2024dreammat, wang2024prolificdreamer} use techniques like SDS to bridge the gap between dimensions, but such methods are not applicable to our deterministic model, and suffer from slow computational speed and mode seeking issues. Therefore, we explore using backprojection to transfer 2D decomposition results from image space to UV space, enabling their application to 3D models. Notably, both SuperMat and SuperMatMV are finite-view models, meaning that one inference can only decompose the visible region from a specific set of viewpoints. To address this problem, we propose estimating materials from multiple viewpoints and integrating them. With multi-view decomposition results from SuperMatMV, the UV refinement network aims to paint missing areas and improve the quality of the generated materials.

\noindent\textbf{Backprojection and UV Blending.} For each viewpoint, we backproject SuperMatMV's decomposition results from image space onto the model's UV texture map, obtaining the partial albedo UV map $C_{d}^i$, RM UV map $C_{rm}^i$, and mask UV map $M^i$ for viewpoint $i$. We then blend the backprojection results from all viewpoints in an averaged manner. For albedo, this can be formulated as:
\begin{equation}
\begin{split}
    M = \sum_{i} M^i, \\
    C_{d} = \frac{\sum_{i} C_{d}^i}{(M + \epsilon)}.
\end{split}
\end{equation}
\noindent
The RM follows the same computation method. The blended partial maps $C_{d}$, $C_{rm}$, and the partial mask $M$ are then input into the UV refinement one-step model.

\noindent\textbf{UV Refinement One-Step Model.} 
Due to the limited number of viewpoints, the blended partial maps often have missing areas. To address this issue, we leverage diffusion models for UV refinement: we fine-tune Stable Diffusion into a UV refinement model, which takes a flawed partial material UV map, a mask UV map, and a position UV map as inputs, and outputs completed and refined material UV maps. Specifically, the material maps are encoded by a VAE and concatenated with downsampled mask and position maps, and then processed by UNet with an expanded 8-channel \textit{conv\_in}. We also modify the model's scheduler to support single-step inference and enable end-to-end training, resulting in a deterministic one-step model.

\noindent\textbf{Material Decomposition Pipeline for 3D Objects.} 
As shown in Fig.~\ref{fig:4}, with the fine-tuned UV refinement network, we propose a complete pipeline for the material decomposition of 3D objects. Given an object with only RGB texture, we first render it from six orthogonal viewpoints. The rendered results are then batch-processed by SuperMatMV for viewpoint-consistent material decomposition. These materials, including albedo and RM, are backprojected into UV space and blended. The blended partial maps are then refined by the UV refinement one-step model and the final material UV maps are generated. This process requires only three inference steps—one for SuperMatMV, one for refining albedo, and one for refining RM—allowing our pipeline to complete the material decomposition of a 3D object in just 3 seconds.

%% file: sec/4_exper.tex
\section{Experiments}

\begin{figure*}
    \centering
    \includegraphics[width=1\linewidth]{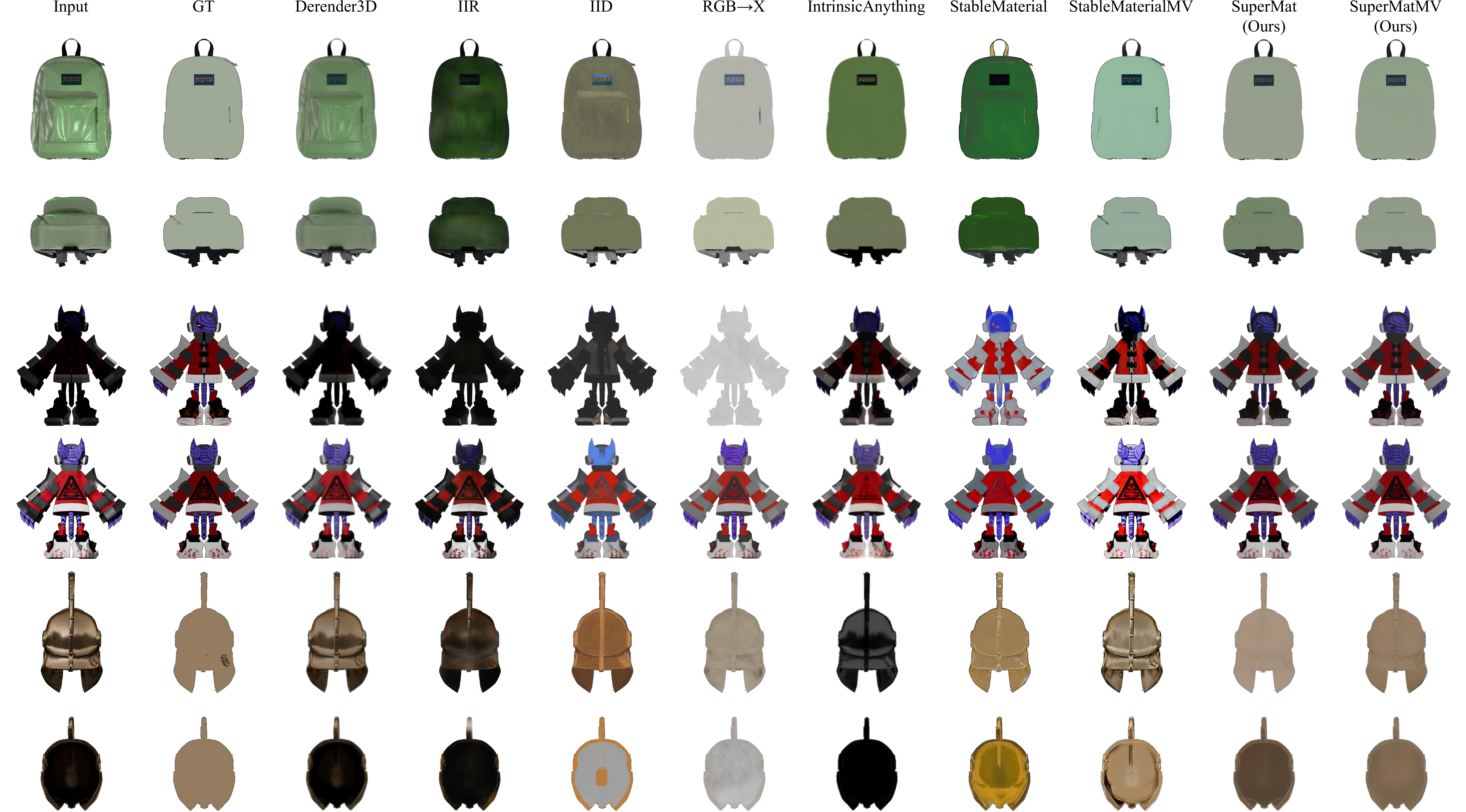}
    \caption{Comparison of our method with others on albedo estimation results. Each pair of rows presents results of the same object under identical lighting conditions from different viewpoints. This organizational format is also followed in other qualitative comparison figures.}
    \label{fig:2}
    \vspace{-10pt}
\end{figure*}

\begin{figure}
    \centering
    \includegraphics[width=1\linewidth]{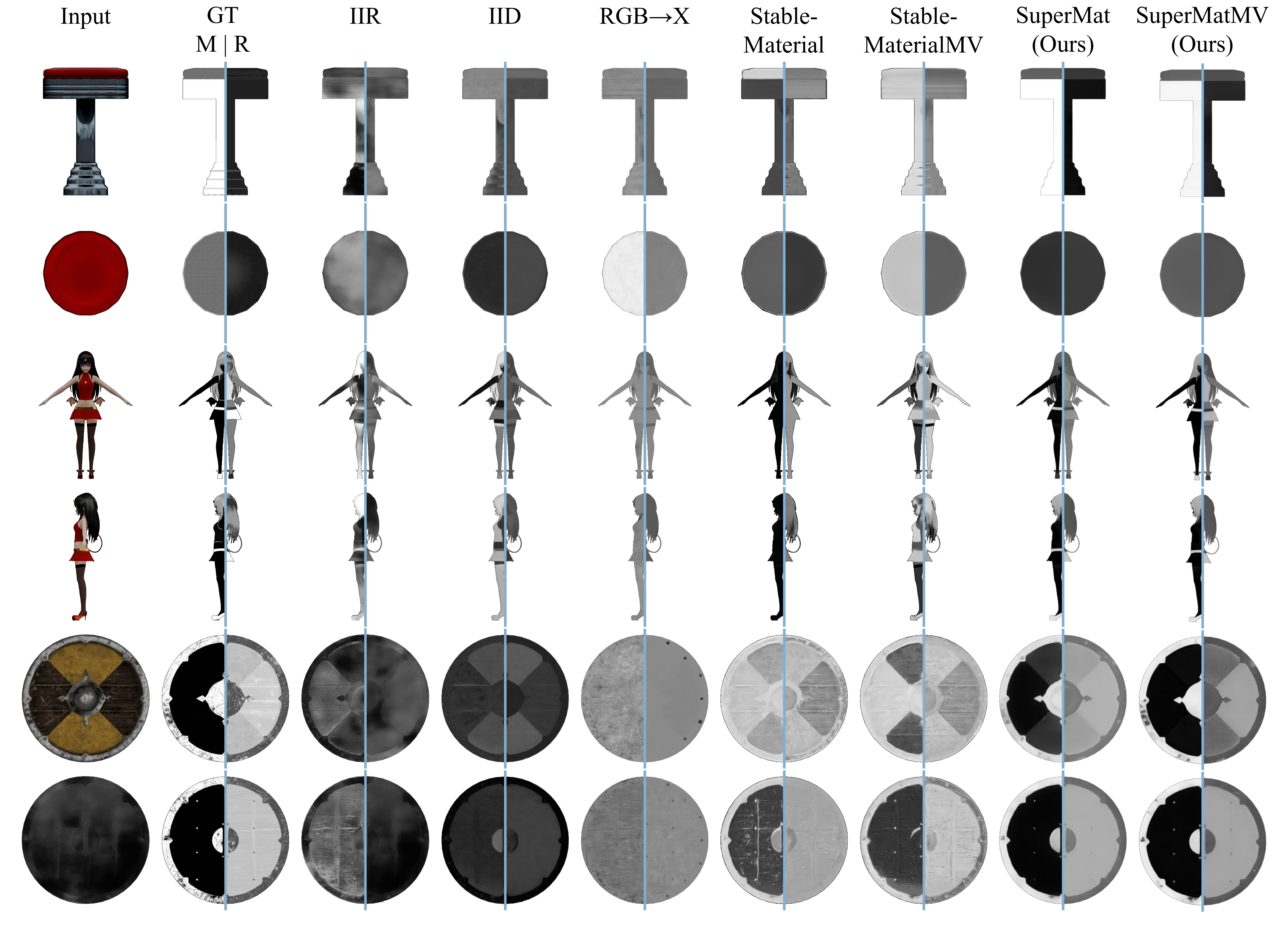}
    \caption{Comparison of our method with others on metallic and roughness estimation results. We combined both materials into a single image, with the left half representing metallic and the right half showing roughness. It can be observed that these two materials exhibit more variation and are more challenging to estimate compared to albedo.}
    \label{fig:1}
    \vspace{-20pt}
\end{figure}

\noindent
\textbf{Training Datasets.} In this paper, we collect 3D objects from Objaverse \cite{deitke2023objaverse} and Digital Twin Catalog (DTC) \cite{dtc} datasets. For each object, we render material maps, normal maps, and position maps from 6 orthogonal viewpoints and generate PBR results under 16 different environment lighting conditions from the same viewpoints. We then filter high-quality data using methods such as DINO score considerations \cite{oquab2023dinov2}, resulting in a dataset of 75k objects that combines both synthetic and real-captured data for the image space decomposition task. Following the approach outlined in Sec.~\ref{sec:3.3}, we further refine this data for the UV refinement network by selecting cases where the missing area in blended partial UV maps is under $50\%$. This process results in a final dataset of 40k objects for training the refinement one-step model.

\noindent\textbf{Quantitative Comparison of Image Space Decomposition.} We randomly select 64 unseen items each from the Objaverse, DTC and BlenderVault \cite{litman2024materialfusion} datasets. Objaverse and BlenderVault primarily contain synthetic data, whereas DTC consists of scanned objects, offering a source of real-world data. Our method is tested against 7 baseline approaches on these 192 objects. For each object, the model processes 18 inputs corresponding to 3 lighting conditions and 6 viewpoints, predicting 3 (or 1) types of materials and generating 54 (or 18) texture maps in total. All outputs are used to compute metrics, with the final averaged results shown in Tab.\ref{tab:1}. We use PSNR, SSIM, and LPIPS metrics to evaluate albedo, metallic, and roughness. Additionally, for methods that provide all material types, we compute metrics on the rendering under relighting conditions using the ground truth normals and positions and the predicted values. The results demonstrate that our approach significantly outperforms the SOTA in image space decomposition.

Beyond image quality, we compare inference speeds by calculating the total time required for each method to process the test set, normalized by the dataset size and number of predicted material types. All tests are conducted with a batch size of 1. For consistency in speed comparison, IID, StableMaterial and StableMaterialMV, which typically average over 10 predictions per input, are modified to perform a single prediction. The “Time(s)” column in Tab.\ref{tab:1} reflects the time required by each method to generate a single material prediction on an NVIDIA A100 GPU with a batch size of 1. Traditional, non-diffusion-based methods like Derender3D and IIR generally show faster speeds. While StableMaterial attempts to predict multiple materials simultaneously by adding extra channels to \textit{conv\_in} and \textit{conv\_out}, its lack of structural design adjustments limits prediction quality and requires multi-step denoising, resulting in slower performance than traditional methods. In contrast, our approach achieves high-quality material decomposition with speeds comparable to traditional methods.

\begin{figure}
    \centering
    \includegraphics[width=1\linewidth]{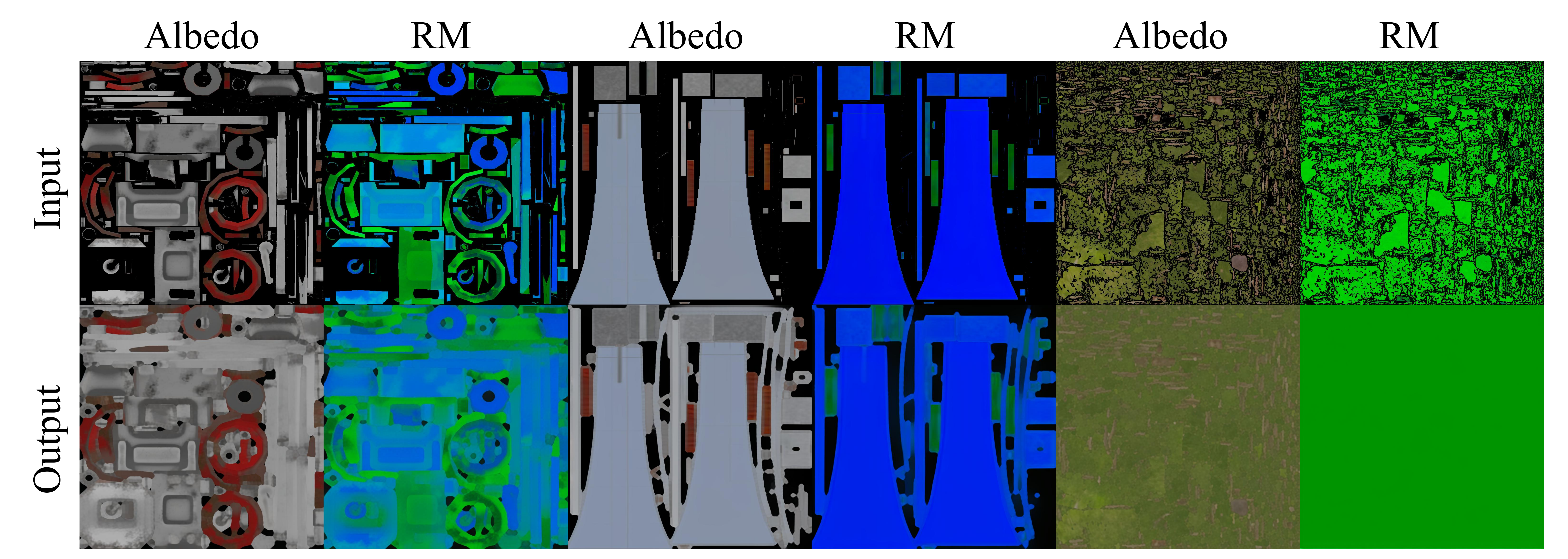}
    \caption{Results of UV refinement one-step model.}
    \label{fig:5}
    \vspace{-20pt}
\end{figure}

\begin{figure*}
    \centering
    \includegraphics[width=1\linewidth]{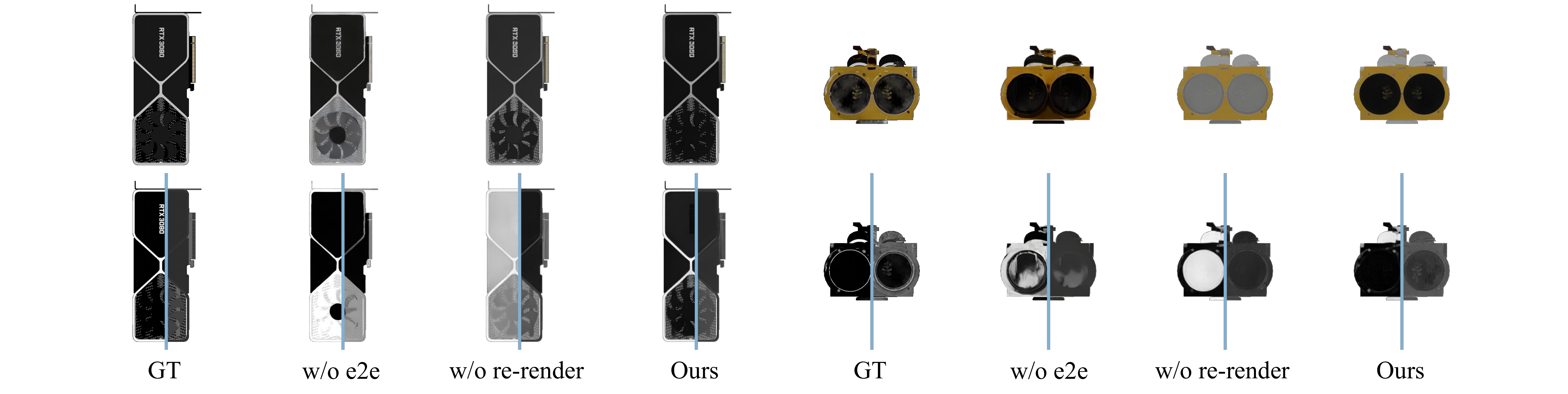}
    \caption{Qualitative ablation results for SuperMat. The first row shows albedo, while the second row shows metallic and roughness.}
    \label{fig:8}
    \vspace{-10pt}
\end{figure*}

\begin{figure}
    \centering
    \includegraphics[width=1\linewidth]{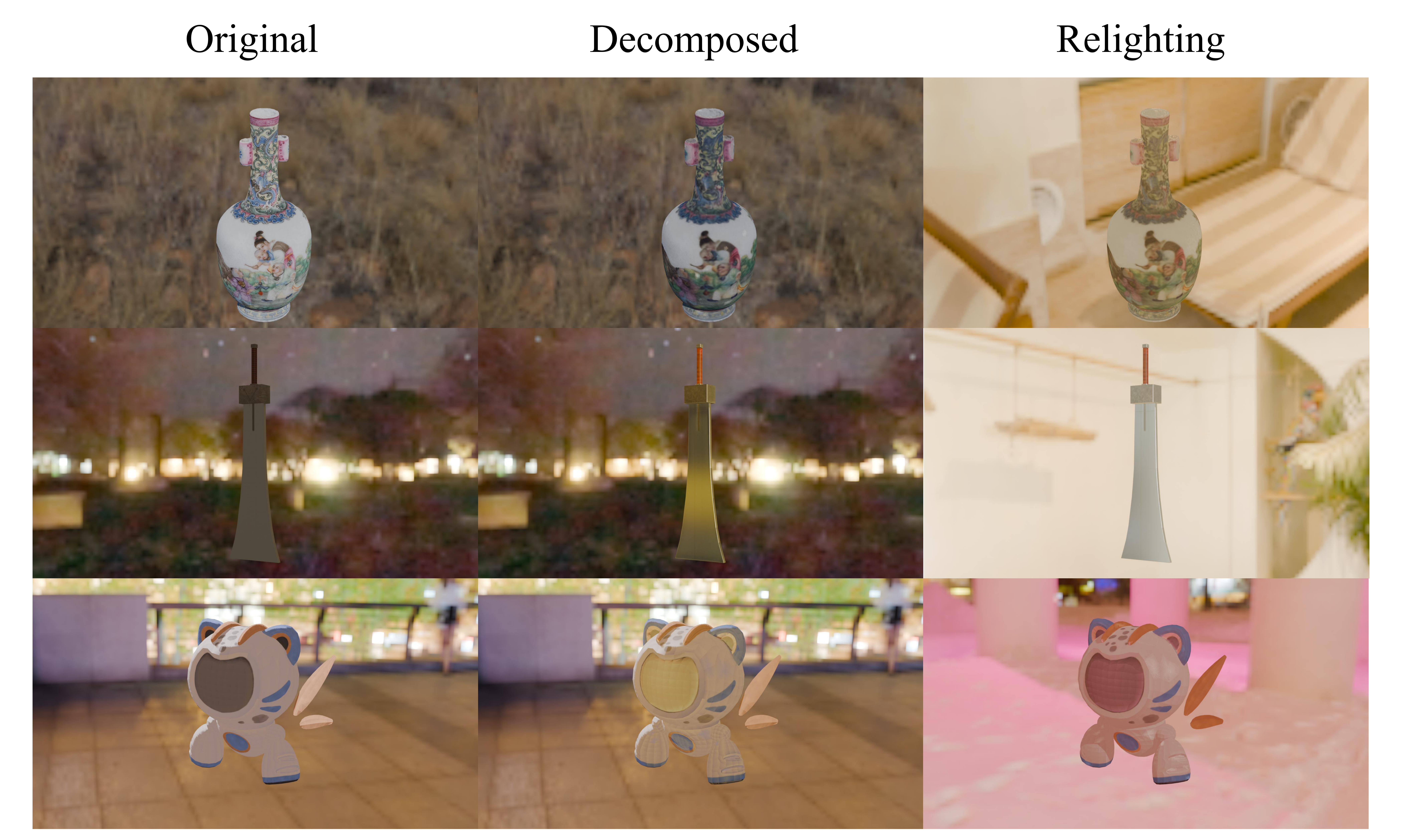}
    \vspace{-15pt}
    \caption{Results of decomposition for 3D objects. We compare the original and decomposed models under the same illumination and show the relighting outcomes of the new models.}
    \label{fig:6}
    \vspace{-15pt}
\end{figure}

\noindent\textbf{Qualitative Comparison of Image Space Decomposition.} Fig.\ref{fig:2} compares albedo prediction results across various cases, including artist-designed objects and real-world scans, showcasing the performance of our method against other approaches. SuperMat demonstrates superior ability in removing highlights and shadows, accurately capturing the intrinsic color, even on challenging metallic surfaces. In Fig.\ref{fig:1}, we present metallic and roughness predictions from our method alongside other baselines. SuperMat outperforms the others with well-defined boundaries between metallic and non-metallic regions and achieves consistency within material types while preserving high-frequency details. Moreover, our results closely align with physical realism, addressing a significant challenge that previous methods struggled to overcome. Additionally, SuperMatMV demonstrates stronger consistency across different views in all material predictions, and is still able to generate reasonable decomposition results for inputs from challenging viewpoints with sparse relevant information. More results will be presented in the supplementary materials.

\noindent\textbf{Results of UV Refinement and 
Material Estimation for 3D Objects.} Fig.\ref{fig:5} illustrates the effectiveness of the UV refinement one-step model in inpainting missing areas in SuperMatMV’s outputs. Fig.\ref{fig:6} displays the results from our material estimation pipeline for 3D models. The first column shows the rendering of the input models with only RGB texture, where all surfaces exhibit uniform reflectance behavior. After applying our method, the decomposed materials enable physically-based rendering with surface-specific reflectance properties, as demonstrated in the second and third columns under different lighting conditions.

\noindent\textbf{Ablation Studies.} In Tab.\ref{tab:1}, we present ablation studies on two core components of SuperMat: end-to-end (e2e) training and re-render loss. E2E with perceptual loss significantly enhances model performance and efficiency. In fact, within the e2e framework, we experiment with three types of loss functions, with perceptual loss achieving the best results. This suggests that direct supervision of the fully denoised predicted material maps, rather than latent noise, plays a critical role in enhancing the model's capability for this specific task. Detailed findings are available in the supplementary materials. The addition of re-render loss, facilitated by the e2e framework, further improves prediction quality, particularly under relighting conditions. Qualitative results of these ablations are shown in Fig. \ref{fig:8}. The ‘w/o e2e’ setting, where the model is trained without perceptual and re-render losses, limits its capacity to remove lighting effects effectively. In contrast, training ‘w/o re-render’ loss leads to minor inaccuracies in fine material details. 

For the decomposition pipeline for 3D objects, we conduct additional ablations on the UV refinement network. Due to limitations from only six orthogonal views, achieving complete texture coverage, accurate merging of overlapping regions, and consistent material transitions are challenging. As shown in Fig.\ref{fig:9}, omitting the UV refinement leads to incomplete textures, visible color-separation lines, and unnatural transitions between materials, underscoring the importance of this component.

%% file: sec/5_conc.tex
\section{Conclusions}

\begin{figure}
    \centering
    \includegraphics[width=1\linewidth]{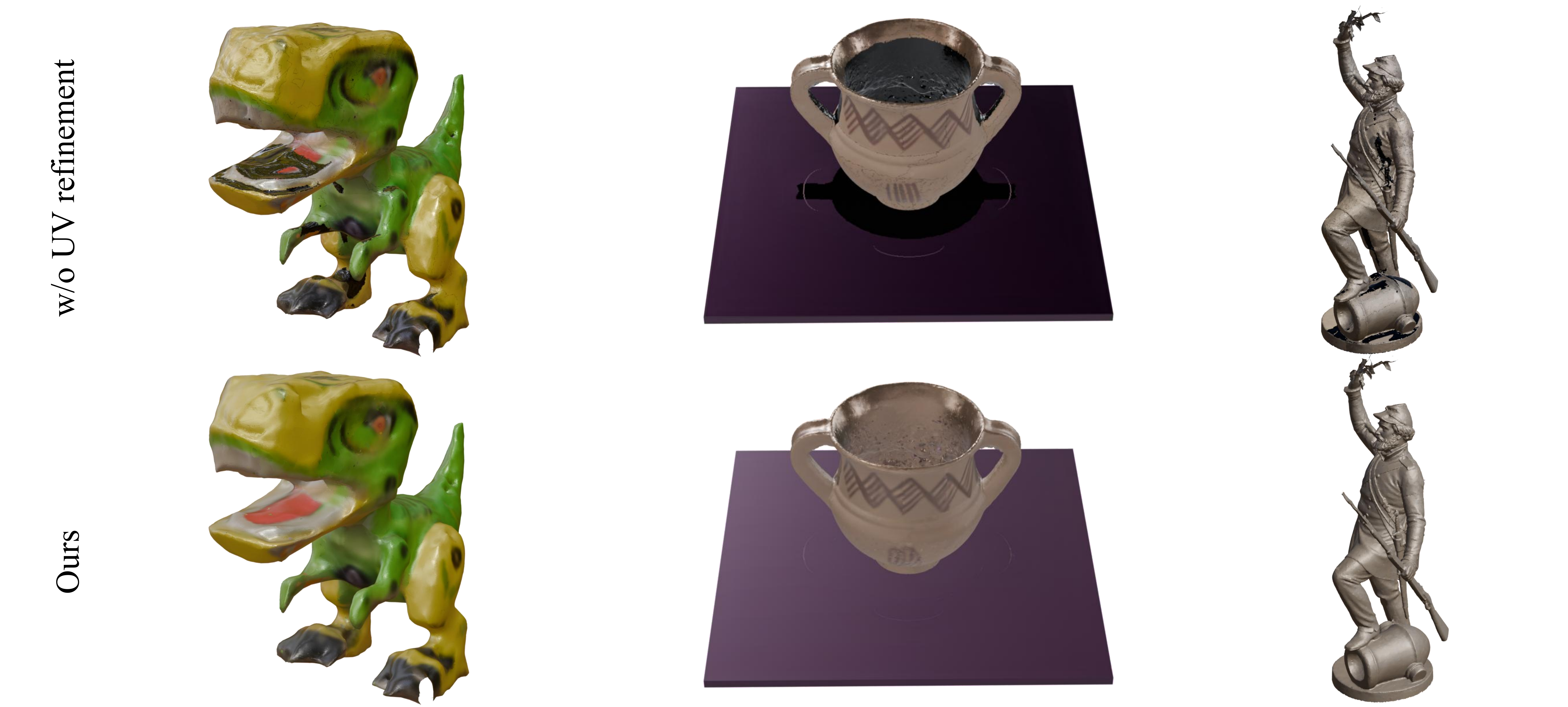}
    \caption{Qualitative ablation results on UV refinement network.}
    \label{fig:9}
    \vspace{-15pt}
\end{figure}

In this paper, we present SuperMat, an innovative image space material decomposition one-step model fine-tuned from Stable Diffusion with task-specific losses in an end-to-end framework. SuperMat enables simultaneous multi-material prediction and supports single-step inference, achieving state-of-the-art quality and outperforming previous diffusion-based methods in efficiency. Through the proposed UV refinement network, we extend SuperMat to decomposition tasks for 3D objects, where it demonstrates exceptional performance and applicability.

%% file: sec/X_suppl.tex
\clearpage
\setcounter{page}{1}
\maketitlesupplementary

\begin{table*}[ht]
\renewcommand{\tabcolsep}{2pt}
\small
\centering
\begin{tabular}{lcccccccccccc}
\toprule
                & \multicolumn{3}{c}{Albedo} & \multicolumn{3}{c}{Metallic} & \multicolumn{3}{c}{Roughness} & \multicolumn{3}{c}{Relighting} \\ \cline{2-13} 
                & PSNR↑    & SSIM↑  & LPIPS↓ & PSNR↑    & SSIM↑   & LPIPS↓  & PSNR↑     & SSIM↑   & LPIPS↓  & PSNR↑     & SSIM↑    & LPIPS↓  \\ \hline
L1              & 26.8102  & \colorbox[HTML]{ffaeb0}{0.9164} & 0.0956 & 22.4801  & \colorbox[HTML]{ffaeb0}{0.8889}  & 0.1803  & 23.5855   & 0.9109  & 0.1194  & 26.8610   & \colorbox[HTML]{ffaeb0}{0.9374}   & 0.0672  \\
SSI             & 24.2359  & 0.9063 & 0.1033 & 22.2521  & 0.8701  & 0.1814  & 23.7918  & 0.9139  & 0.1165  & 24.7511   & 0.9266   & 0.0732  \\
Perceptual & \colorbox[HTML]{ffaeb0}{27.0082} & 0.9151 & \colorbox[HTML]{ffaeb0}{0.0949} & \colorbox[HTML]{ffaeb0}{22.8702} & 0.8669 & \colorbox[HTML]{ffaeb0}{0.1760} & \colorbox[HTML]{ffaeb0}{24.1452} & \colorbox[HTML]{ffaeb0}{0.9145} & \colorbox[HTML]{ffaeb0}{0.1156} & \colorbox[HTML]{ffaeb0}{27.2484} & \colorbox[HTML]{ffaeb0}{0.9374} & \colorbox[HTML]{ffaeb0}{0.0650} \\
\bottomrule
\end{tabular}
\caption{Quantitative comparison across different loss functions. We highlight the \colorbox[HTML]{ffaeb0}{best} results for each metric.}
\label{tab:6}
\end{table*}

\begin{table*}[ht]
\renewcommand{\tabcolsep}{2pt}
\small
\centering
\begin{tabular}{llccclccclccc}
\toprule
\multirow{2}{*}{Training Data Type} & \multicolumn{4}{c}{Albedo} & \multicolumn{4}{c}{Metallic} & \multicolumn{4}{c}{Roughness} \\ \cline{2-13} 
                                    & MSE↓   & PSNR↑  & SSIM↑  & LPIPS↓  & MSE↓ & PSNR↑  & SSIM↑ & LPIPS↓  & MSE↓  & PSNR↑ & SSIM↑                      & LPIPS↓                     \\ \hline
Synthetic                           & 0.0027 & 27.41                     & 0.9195                     & 0.0885                     & 0.0120                                          & \colorbox[HTML]{ffaeb0}{23.80}                     & 0.8767                     & \colorbox[HTML]{ffaeb0}{0.1692}                     & 0.0084                                          & 24.32                     & 0.9152                     & 0.1118                     \\
Real-captured                       &   0.0061     &     24.50                        &       0.8982                &        0.1049         &                           0.0350                      &      19.40                       &           0.8464                 &        0.2043                    &                0.0113                                 &      22.32                       &         0.9075                   &      0.1165                      \\
Synthetic+Real-captured           & \colorbox[HTML]{ffaeb0}{0.0025} & \colorbox[HTML]{ffaeb0}{27.58} & \colorbox[HTML]{ffaeb0}{0.9205} & \colorbox[HTML]{ffaeb0}{0.0867} & \colorbox[HTML]{ffaeb0}{0.0111}                                          & 23.77 & \colorbox[HTML]{ffaeb0}{0.8787} & 0.1696 & 0.0074                                          & \colorbox[HTML]{ffaeb0}{24.63} & \colorbox[HTML]{ffaeb0}{0.9154} & \colorbox[HTML]{ffaeb0}{0.1115} \\
\bottomrule
\end{tabular}
\caption{Ablation results on training data types in the image space decomposition task. We highlight the \colorbox[HTML]{ffaeb0}{best} results for each metric.}
\label{tab:9}
\end{table*}

\begin{table}[!ht]
\renewcommand{\tabcolsep}{2pt}
\small
\centering
\begin{tabular}{llcllcl}
\toprule
        & \multicolumn{3}{c}{Albedo}                                            & \multicolumn{3}{c}{RM}                                                \\ \cline{2-7} 
        & \multicolumn{1}{c}{PSNR↑} & SSIM↑ & \multicolumn{1}{c}{LPIPS↓} & \multicolumn{1}{c}{PSNR↑} & SSIM↑ & \multicolumn{1}{c}{LPIPS↓} \\ \hline
Input   & 13.56     &    0.5499                       &   0.4227     &      12.52 &      0.5484                     &   0.5245     \\
Refined &   23.92    &  0.7537                         &   0.2792     &     27.31  &       0.8460                    &  0.1178     \\
\bottomrule
\end{tabular}
\caption{Quantitative evaluation on UV maps before and after refinement.}
\label{tab:7}
\end{table}

\begin{table}[ht]
\setlength{\tabcolsep}{0pt}
\small
\centering
\begin{tabular}{lcccccc}
\toprule
                  & \multicolumn{3}{c}{BlenderVault} & \multicolumn{3}{c}{StanfordORB} \\ \cline{2-7} 
                  & PSNR↑     & SSIM↑    & LPIPS↓    & PSNR↑    & SSIM↑    & LPIPS↓    \\ \hline
Derender3D        &   22.52        &   0.9151       &   0.1023        &    17.03      &   0.8241       &    0.2012       \\
IIR               &   21.10        &   0.9005       &  0.1046         &    15.83      &   0.7944       &    0.2005       \\
IID               &  22.10         &  0.9095        &   0.0989        &    16.21      &   0.8000       &    0.2059       \\
RGB→X             &  22.23         &   0.9115       &   0.0987        &    \colorbox[HTML]{ffffa8}{17.23}      &   \colorbox[HTML]{ffffa8}{0.8272}      &    0.1849       \\
IA &   23.33        &  0.9115        &   0.1047        &    \colorbox[HTML]{ffd7ac}{17.98}      &   \colorbox[HTML]{ffd7ac}{0.8286}       &    0.1875       \\
SM    &   \colorbox[HTML]{ffffa8}{23.36}        &  \colorbox[HTML]{ffd7ac}{0.9176}        &  \colorbox[HTML]{ffffa8}{0.0909}         &    16.94      &   0.8174       &    \colorbox[HTML]{ffffa8}{0.1801}       \\
SMMV  &   \colorbox[HTML]{ffd7ac}{23.57}        & \colorbox[HTML]{ffffa8}{0.9175}         &   \colorbox[HTML]{ffd7ac}{0.0887}        &    17.10      &   0.8189       &    \colorbox[HTML]{ffd7ac}{0.1785}       \\
Ours              &  \colorbox[HTML]{ffaeb0}{27.00}         &  \colorbox[HTML]{ffaeb0}{0.9463}        &   \colorbox[HTML]{ffaeb0}{0.0629}        &  \colorbox[HTML]{ffaeb0}{24.99}        &  \colorbox[HTML]{ffaeb0}{0.9281}        &   \colorbox[HTML]{ffaeb0}{0.0962}       
 \\
\bottomrule
\end{tabular}
\caption{Quantitative comparison of novel view synthesis relighting. We highlight the \colorbox[HTML]{ffaeb0}{best}, \colorbox[HTML]{ffd7ac}{second-best}, and \colorbox[HTML]{ffffa8}{third-best} results for each metric. Here “IA", “SM", “SMMV" represents “IntrinsicAnything", “StableMaterial", “StableMaterialMV" respectively.}
\label{tab:8}
\end{table}

\section{Physically Based Rendering (PBR) Model}

We use the Cook-Torrance Bidirectional Reflectance Distribution Function (BRDF) \cite{cook1982reflectance} based on microfacet theory to define the materials and establish the rendering model. Following \cite{chen2023fantasia3d}, for a point with coordinates $p \in \mathbb{R}^3$, albedo $a \in \mathbb{R}^3$, metallic $m \in \mathbb{R}$, roughness $r \in \mathbb{R}$, and surface orientation $n \in \mathbb{R}^3$, the PBR result $L$ observed from viewpoint $c \in \mathbb{R}^3$ is given by:
\begin{equation}
\label{eq:brdf}
\begin{split}
    L(p, \omega) = a (1 - m) \int_\Omega L_i(p, \omega_i)(\omega_i \cdot n) \mathrm{d} \omega_i \\ 
    + \int_\Omega \frac{DFG}{4(\omega \cdot n)(\omega_i, n)} L_i(p, \omega_i)(\omega_i \cdot n) \mathrm{d} \omega_i,
\end{split}
\end{equation}

\noindent
where $\omega$ represents the direction of the outgoing light from point $p$ to $c$, \emph{i.e.}, the viewing direction. $L_i$ denotes the incident light from the direction $\omega_i$, and $\Omega = \{\omega_i: \omega_i \cdot n \geq 0\}$ represents the hemisphere of normals. $D$, $F$, and $G$ are the distribution, Fresnel, and geometry functions, respectively. For the integral part, the time complexity of Monte Carlo methods is unacceptable. Given that we use ambient lighting as the light source, we compute it efficiently using the split-sum method \cite{munkberg2022extracting}.

\section{Additional Details}

\noindent\textbf{Implementation Details.} We report some implementation details as follows: 1) In the UV refinement one-step model, the input channels are expanded to 8, with the weights of the additional 4 channels initialized to 0. 2) In the ablation experiment, the structure of SuperMat under the ``w/o e2e" setting is slightly different. Without single-step inference, we can only train the diffusion model in a denoising task manner, meaning that during the single-step denoising process, the latents that the UNet receives are not encoded from a clean rendered image, but rather a noisy albedo and noisy RM. Therefore, on top of SuperMat, we additionally replicate the \textit{conv\_in} and the first \textit{DownBlock} as independent parts of structural expert branches to map the inputs from two different domains to similar distributions. These are then fused in an averaged manner before being fed into the shared modules. 3) In the re-render loss implementation, each time we perform relighting, we randomly select a lighting condition from a set of 50 environment maps, covering nearly all possible lighting scenarios.

\noindent
\textbf{Training Details.} Both the SuperMat and the UV refinement model are fine-tuned from Stable Diffusion 2.1, while SuperMatMV is built upon SuperMat. We train SuperMat using the AdamW optimizer with a learning rate of $2e-5$ on 8 NVIDIA A800 (80GB) GPUs, with a batch size of 32, for a total of 30 epochs. The UV refinement one-step model is trained with the AdamW optimizer at a learning rate of $2e-5$, also on 8 NVIDIA A800 (80GB) GPUs, with a batch size of 16, for 40 epochs. The images are resized to resolutions of $512 \times 512$ and $1024 \times 1024$, used for SuperMat and the UV refinement one-step model, respectively. The training setup for SuperMatMV mirrors that of SuperMat, except that SuperMatMV is trained for only 3 epochs using a batch size of 8 for 6-view images.

\noindent
\textbf{Image Space Decomposition Baselines.} We compare SuperMat and SuperMatMV with 7 other image space decomposition networks. Inverse Indoor Rendering (IIR) \cite{zhu2022learning}, Intrinsic Image Diffusion (IID) \cite{kocsis2024intrinsic}, and RGB$\rightarrow$X \cite{zeng2024rgb} are scene-level material estimation methods. Derender3D \cite{wimbauer2022rendering} and IntrinsicAnything \cite{chen2024intrinsicanything} are adaptable to diverse data but do not generate all material types. StableMaterial \cite{litman2024materialfusion} and its multi-view version, StableMaterialMV, provide image space material denoising diffusion priors to MaterialFusion and, like SuperMat, focus on decomposing object materials. It is worth noting that, for scene-targeted methods, to ensure a fairer comparison, we re-train a deterministic method, IIR,  and a diffusion-based method, RGB$\rightarrow$ X, on our dataset.

\section{Additional Visualizations}

\begin{figure*}
    \centering
    \includegraphics[width=1\linewidth]{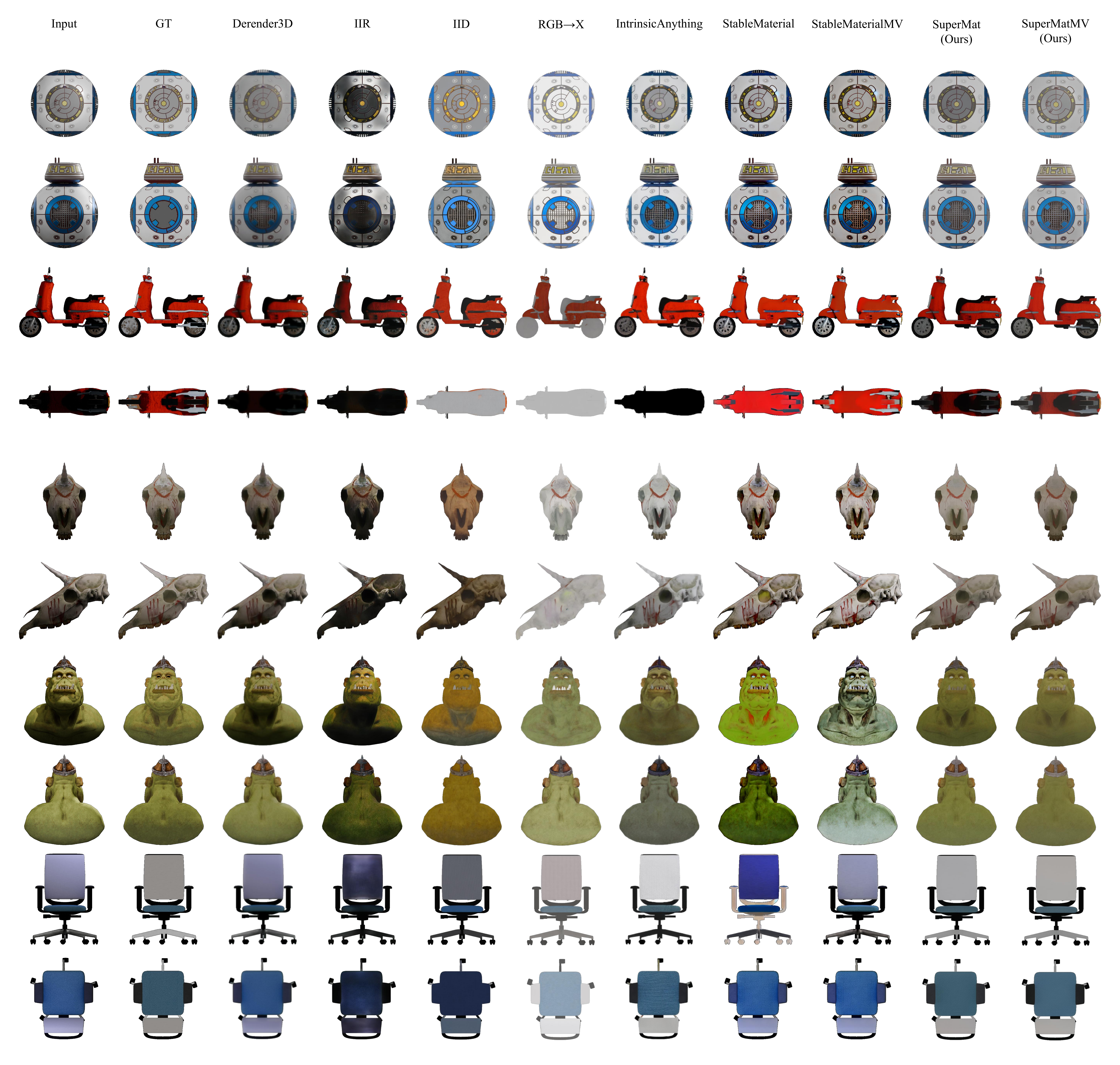}
    \caption{Additional albedo comparison from the Objaverse test dataset.}
    \label{fig:11}
\end{figure*}

\begin{figure*}
    \centering
    \includegraphics[width=1\linewidth]{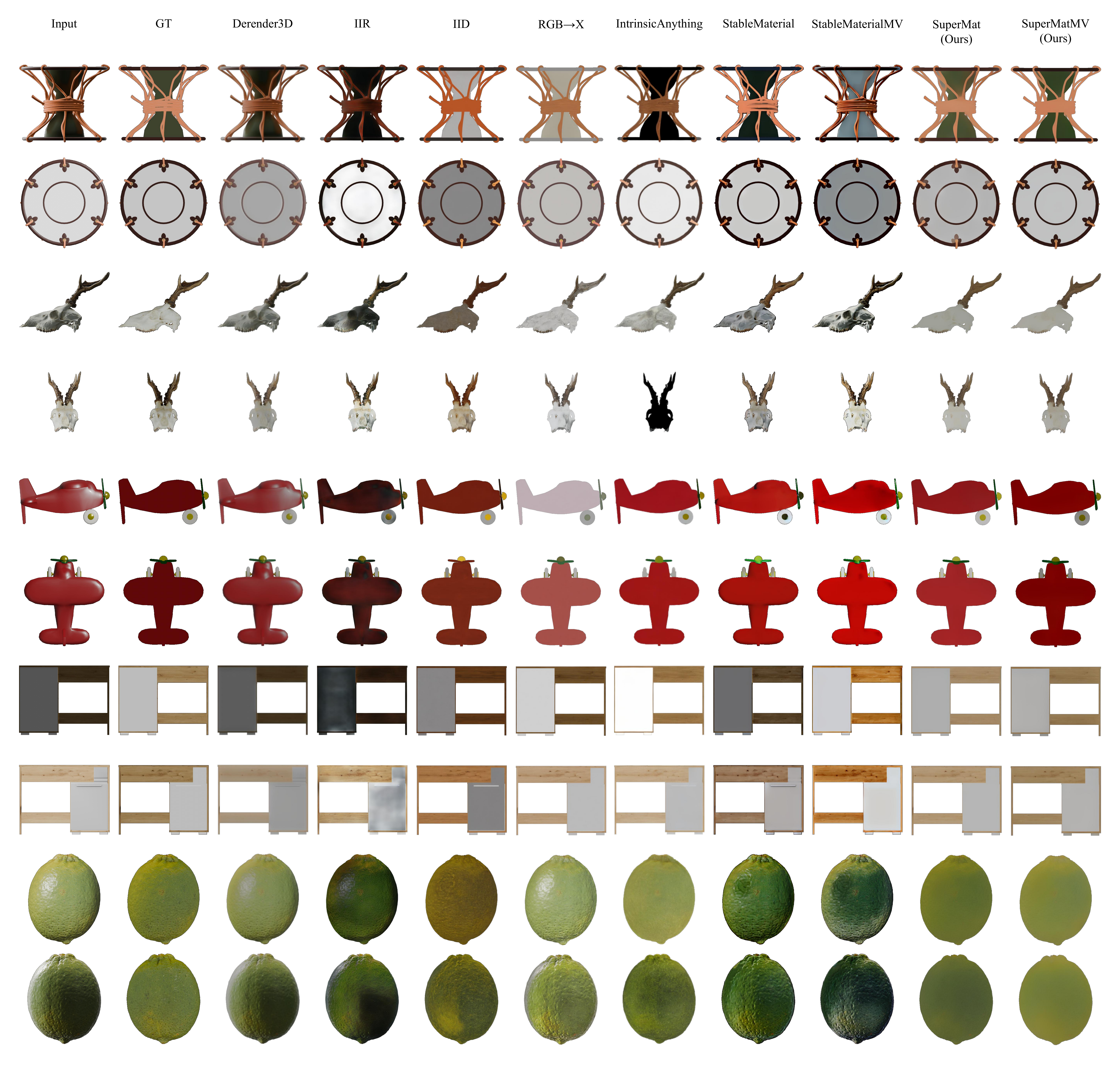}
    \caption{Additional albedo comparison from the BlenderVault test dataset.}
    \label{fig:10}
\end{figure*}

\begin{figure*}
    \centering
    \includegraphics[width=1\linewidth]{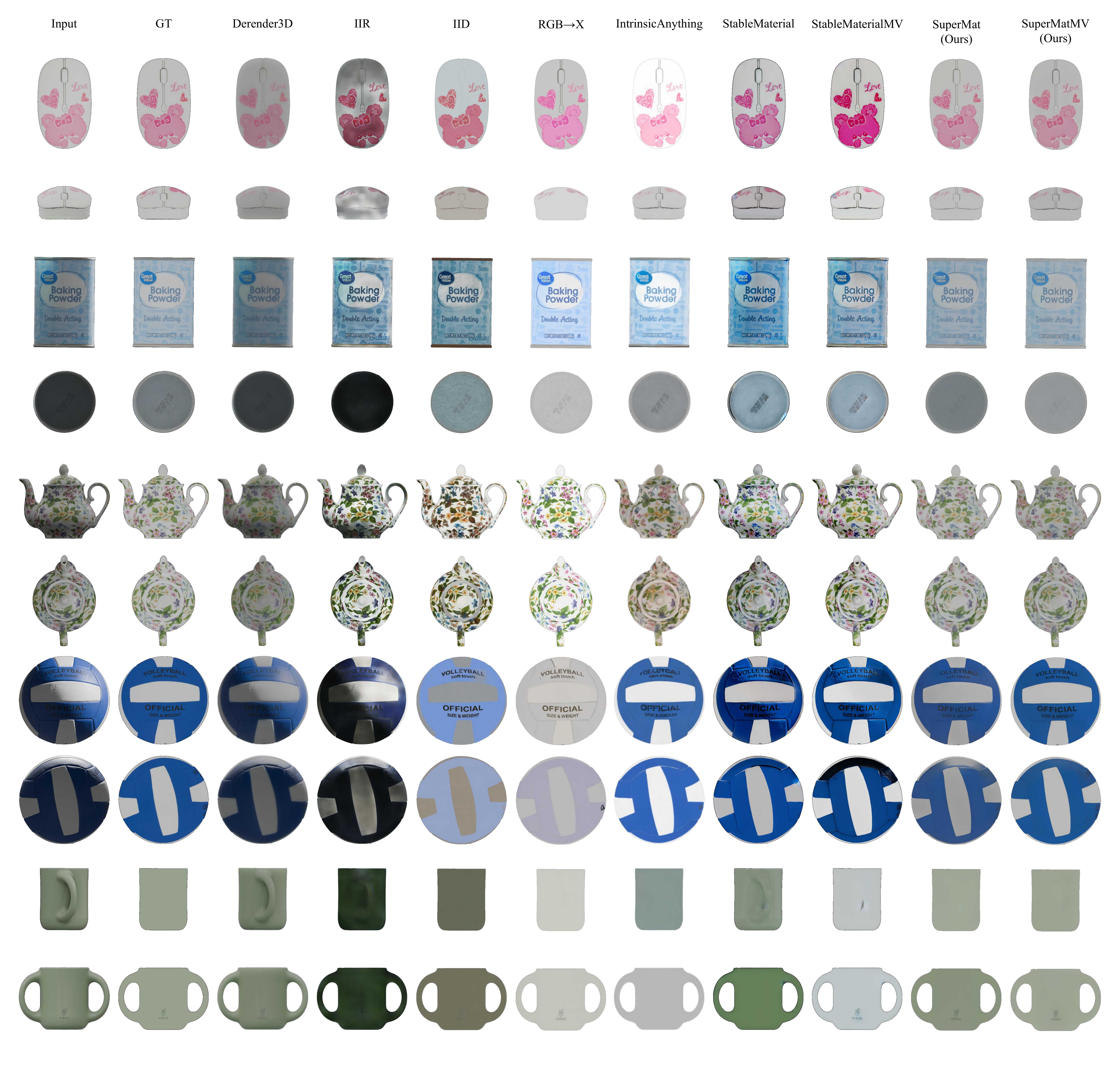}
    \caption{Additional albedo comparison from the DTC test dataset.}
    \label{fig:18}
\end{figure*}

\begin{figure*}
    \centering
    \includegraphics[width=1\linewidth]{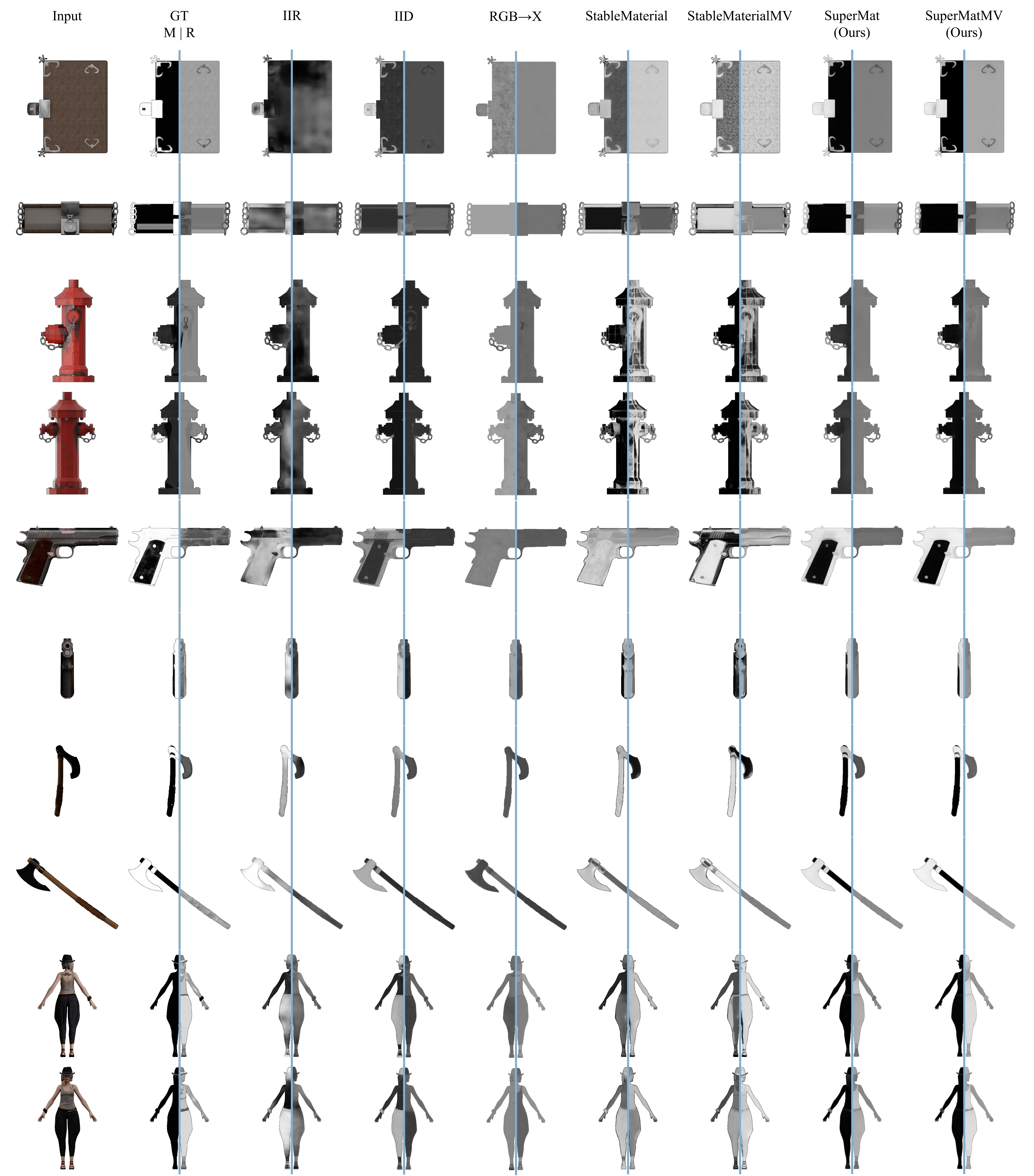}
    \caption{Additional metallic and roughness comparison from the Objaverse test dataset. The metallic maps are shown on the left side (M), while the roughness maps are shown on the right side (R).}
    \label{fig:12}
\end{figure*}

\begin{figure*}
    \centering
    \includegraphics[width=1\linewidth]{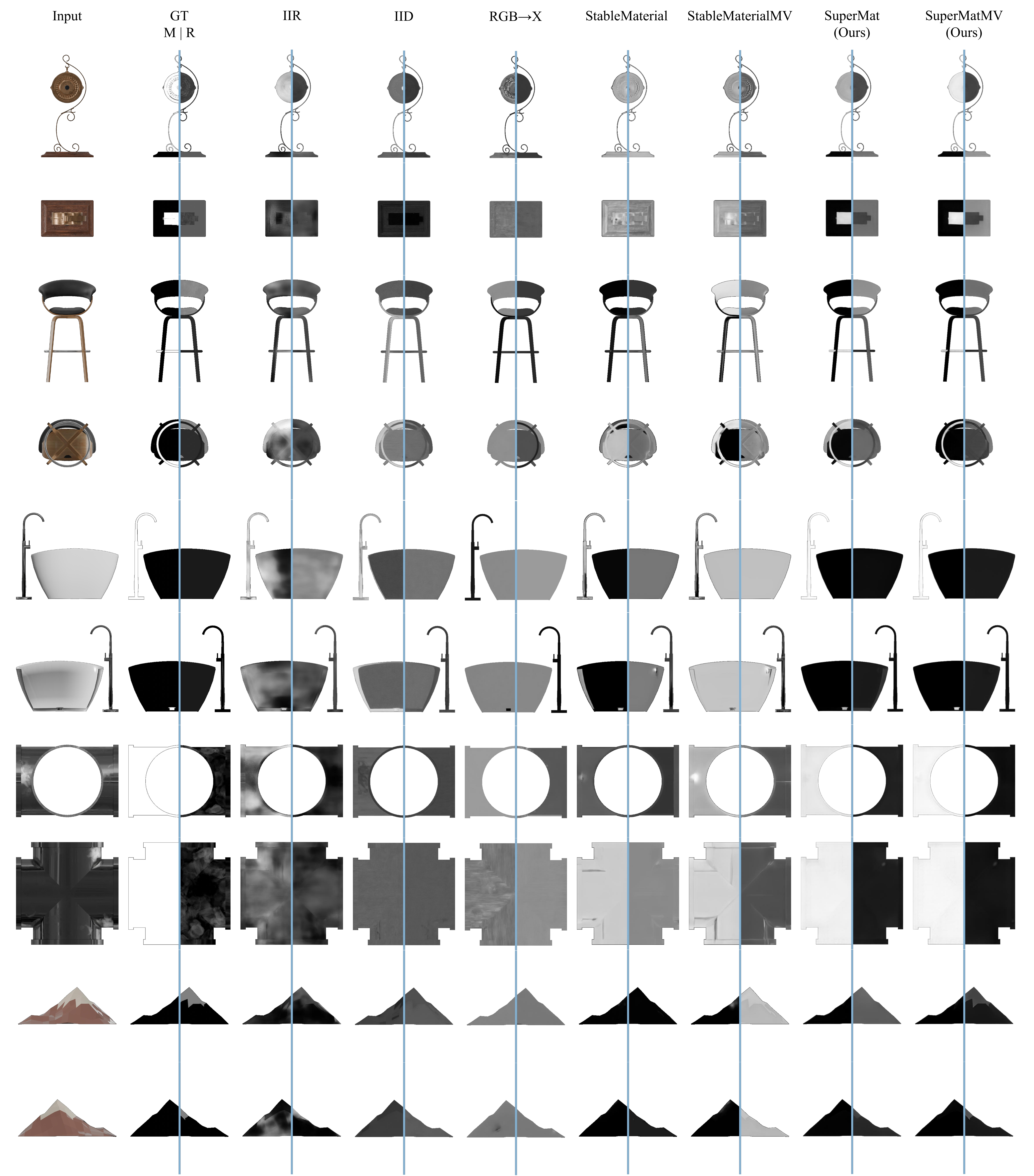}
    \caption{Additional metallic and roughness comparison from the BlenderVault test dataset. The metallic maps are shown on the left side (M), while the roughness maps are shown on the right side (R).}
    \label{fig:13}
\end{figure*}

\begin{figure*}
    \centering
    \includegraphics[width=1\linewidth]{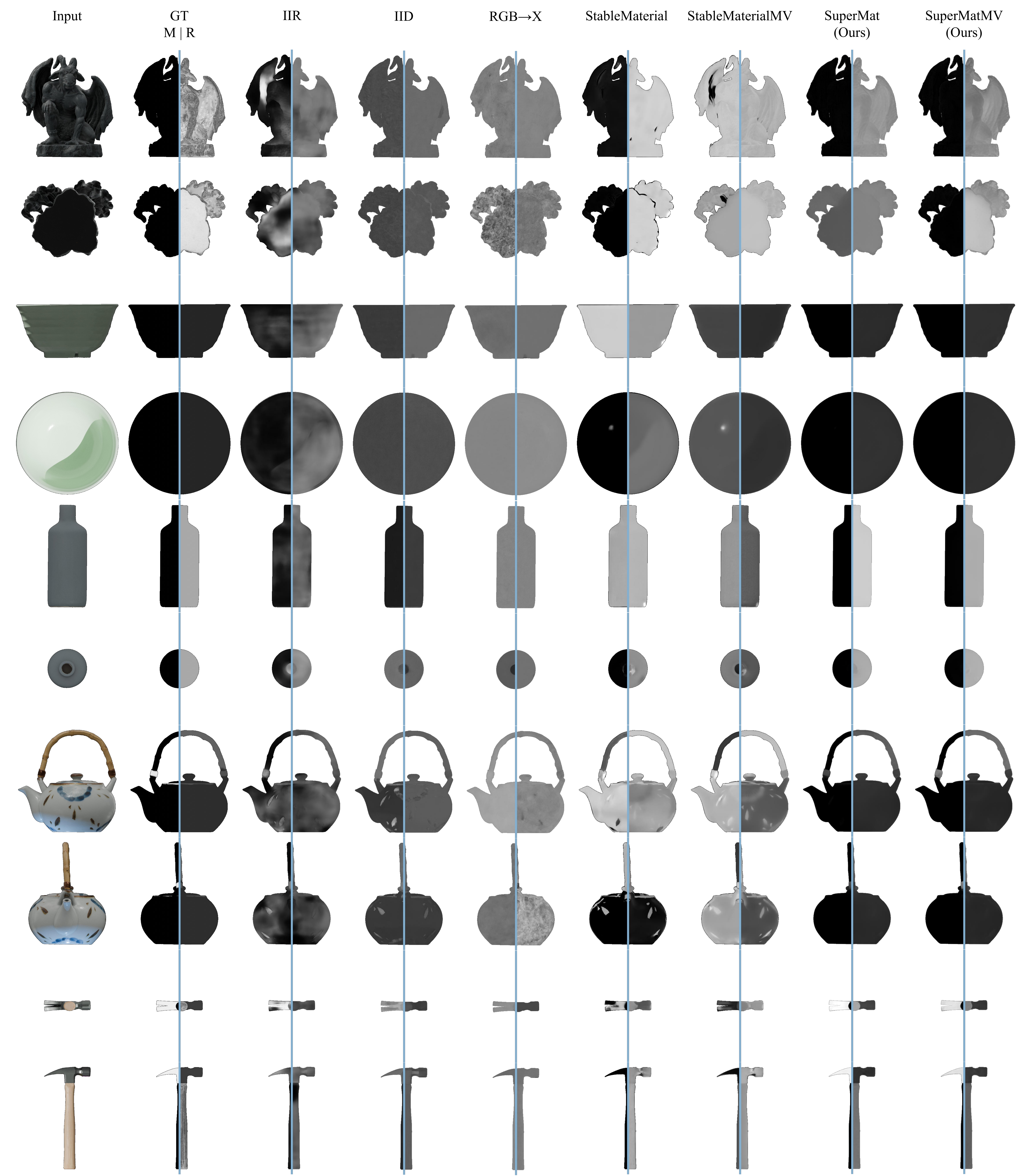}
    \caption{Additional metallic and roughness comparison from the DTC test dataset. The metallic maps are shown on the left side (M), while the roughness maps are shown on the right side (R).}
    \label{fig:19}
\end{figure*}

\begin{figure*}
    \centering
    \includegraphics[width=1\linewidth]{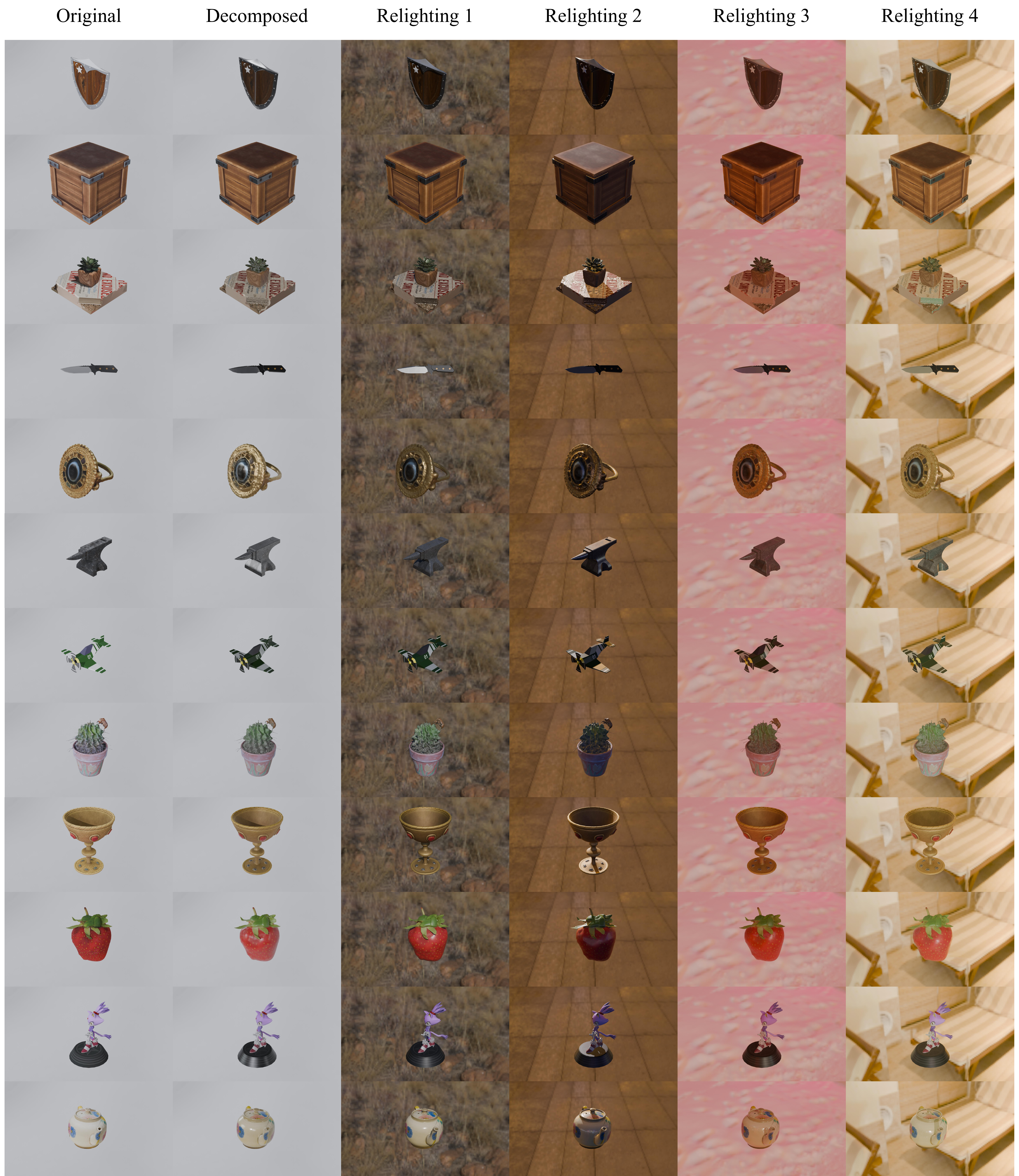}
    \caption{Additional results of decomposition for 3D objects.}
    \label{fig:15}
\end{figure*}

In Figures~\ref{fig:11},~\ref{fig:10},~\ref{fig:18},~\ref{fig:12},~\ref{fig:13},~\ref{fig:19}, we present additional results of SuperMat on the Objaverse \cite{deitke2023objaverse}, BlenderVault \cite{litman2024materialfusion} and DTC \cite{dtc} test datasets, covering both artist-designed and real-world scanned objects. Fig.~\ref{fig:15} illustrates more decomposition results on 3D objects.

\section{Additional Experiments}

\begin{figure*}
    \centering
    \includegraphics[width=1\linewidth]{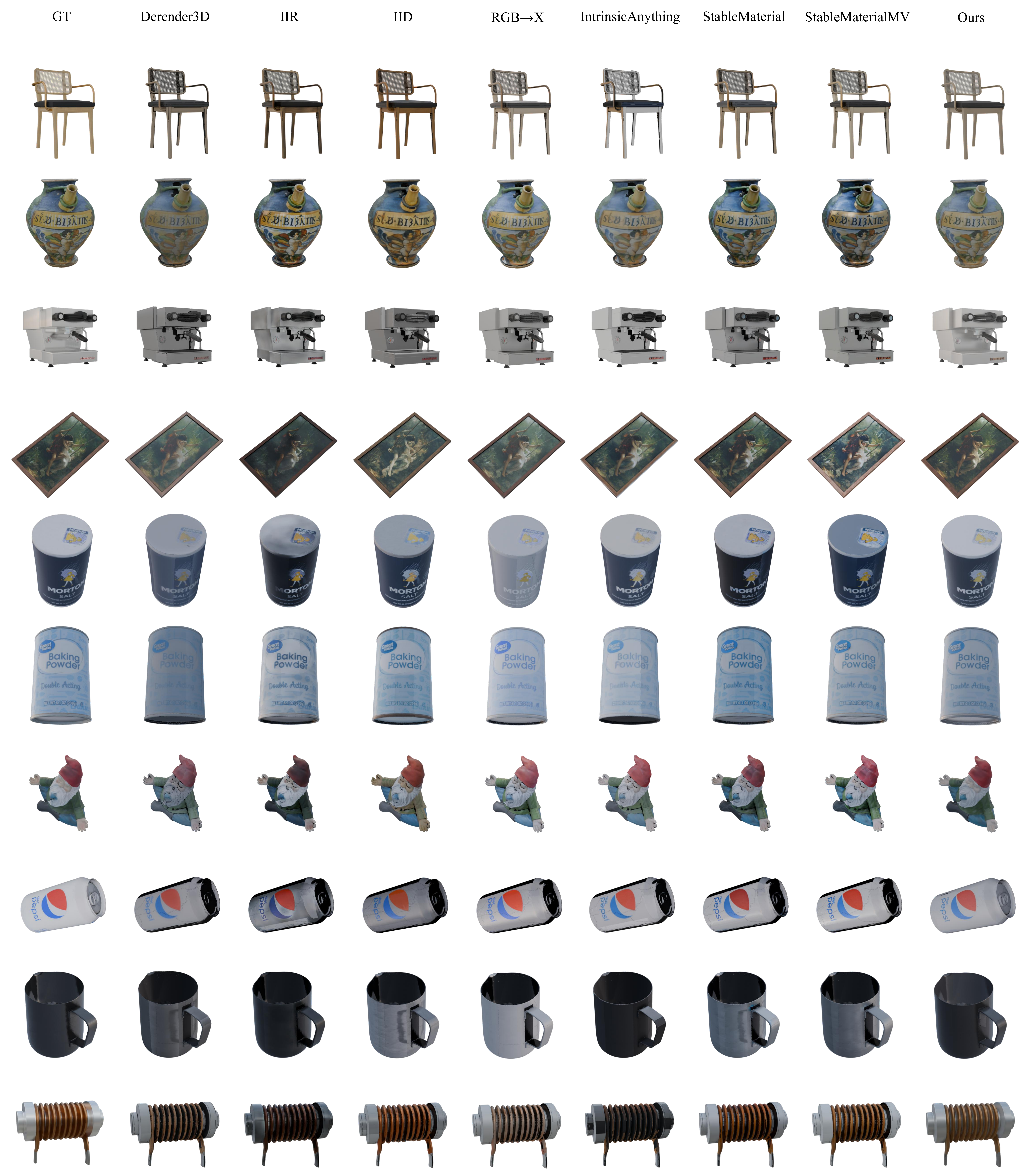}
    \caption{Qualitative comparison of novel view synthesis relighting.}
    \label{fig:20}
\end{figure*}

\textbf{Novel View Synthesis Relighting.} We validate our material decomposition pipeline for 3D objects by relighting the decomposed objects under novel lighting and viewpoints and comparing the results with ground-truth relit object images. We select all 14 objects from the StanfordORB dataset \cite{kuang2023stanfordorb} and 14 randomly chosen unseen objects from the BlendVault dataset \cite{litman2024materialfusion}, applying appropriate lighting conditions before performing material decomposition. We adapt baseline methods to incorporate the same UV backprojection and blending framework as our approach, allowing them to handle 3D object decomposition. We then render the decomposed models from 60 novel viewpoints under three different environment maps. The quantitative comparison results are presented in Tab. \ref{tab:8}, while qualitative results are shown in Fig. \ref{fig:20}. Our method significantly outperforms others in both texture completeness and the physical consistency of materials, demonstrating its effectiveness in decomposing high-quality materials for 3D objects.

\noindent\textbf{Comparison between Different Loss Functions.} Without re-render loss, we experiment with three types of loss functions in the end-to-end framework: L1 loss, shift and scale invariant (SSI) loss \cite{ranftl2020towards}, and perceptual loss. The performance of models trained with these loss designs is evaluated on the same test dataset in Objaverse \cite{deitke2023objaverse}, and the results are shown in Tab.~\ref{tab:6}. Among these losses, the perceptual loss, which is ultimately adopted by SuperMat, demonstrates the best performance. 

\noindent\textbf{Quantitative Validation of UV Refinement.} Tab. \ref{tab:7} presents the quantitative evaluation results of the UV maps generated and blended by SuperMatMV before and after refinement, validating the effectiveness of the UV refinement one-step model.

\noindent\textbf{Ablation on the Training Dataset.} For the image space decomposition task, we conduct an ablation study on the composition of the training dataset, testing the SuperMat's performance when trained solely on synthetic data or only on real-captured data. We randomly select 64 objects containing both data types from the training set, totaling 6144 inputs, which are unseen during training but used to evaluate the model's estimations of 3 material types with MSE, PSNR, SSIM, and LPIPS metrics. The results of the experiment are shown in Tab.\ref{tab:9}, indicating that data diversity helps improve the model's generalization ability.

\noindent\textbf{Detailed Quantitative Comparison.} We show the detailed quantitative comparison of the image space decomposition task in Tables~\ref{tab:2},~\ref{tab:3},~\ref{tab:4} and ~\ref{tab:5}. 

\section{Limitations}

Although our method enables fast and physically consistent material decomposition for both images and 3D objects, it still has several limitations. First, converting the diffusion model into a deterministic model significantly improves performance but also sacrifices the benefits of the diffusion process. For instance, in cases with high-frequency details, SuperMat's results may lack sharpness. Additionally, as a deterministic model, SuperMat has reduced flexibility—once an estimation error occurs, rerunning the process will always yield the same incorrect result. Furthermore, due to the constraints of the chosen BRDF model, SuperMat faces challenges in handling transparent materials, multi-layered surfaces, and highly reflective objects.

\begin{table*}[ht]
\renewcommand{\tabcolsep}{2pt}
\small
\centering
\begin{tabular}{lcccccccccccccccc}
\toprule
\multicolumn{1}{c}{} & \multicolumn{4}{c}{Objaverse}      & \multicolumn{4}{c}{BlenderVault}   & \multicolumn{4}{c}{DTC}\\ \cline{2-13} 
\multicolumn{1}{c}{} & MSE↓   & PSNR↑   & SSIM↑  & LPIPS↓ & MSE↓   & PSNR↑   & SSIM↑  & LPIPS↓ & MSE↓   & PSNR↑   & SSIM↑  & LPIPS↓ \\ \hline
Derender3D           & 0.0087 & 22.16 & 0.8532 & 0.1816 & 0.0104 & 22.45 & 0.8919 & 0.1489 &  0.0246 & 18.30 & 0.8366 & 0.2157 \\
IIR                  & 0.0100 & 22.99 & 0.8741 & 0.1396 & 0.0150 & 22.32 & 0.8709 & 0.1616 & 0.0165 & 20.50 & 0.8677 & 0.1644 \\
IID                  & 0.0088 & 23.15 & 0.8847 & 0.1276 & 0.0122 & 22.39 & 0.8845 & 0.1395 & 0.0101 & 22.15 & 0.9031 & 0.1175 \\
RGB$\rightarrow$X    & 0.0073 & 23.56 & 0.8905 & 0.1017 & 0.0141 & 21.73 & 0.8757 & 0.1332 & 0.0129 & 21.60 & 0.8917 & 0.1114 \\
IntrinsicAnything    & 0.0125 & 21.36 & 0.8701 & 0.1550 & 0.0219 & 20.24 & 0.8700 & 0.1682 & 0.0122 & 21.23 & 0.8963 & 0.1328 \\
StableMaterial       & 0.0110 & 23.78 & 0.8989 & 0.1064 & 0.0143 & 23.90 & 0.9061 & 0.1079 & 0.0129 & 22.64 & 0.9008 & 0.1086 \\ 
StableMaterialMV     & 0.0074 & 24.41 & 0.8974 & 0.0970 & 0.0089 & 25.13 & 0.9071 & \colorbox[HTML]{ffd7ac}{0.1014} & 0.0067 & 24.15 & 0.9229 & 0.0855 \\ \hline
SuperMat w/o e2e     & 0.0057 & 25.11 & 0.8973 & 0.1027 & 0.0102 & 24.30 & 0.8935 & 0.1219 & 0.0107 & 23.36 & 0.8985 & 0.1038 \\
SuperMat w/o re-render   & \colorbox[HTML]{ffffa8}{0.0029} & \colorbox[HTML]{ffffa8}{27.01} & \colorbox[HTML]{ffffa8}{0.9151} & \colorbox[HTML]{ffffa8}{0.0949} & \colorbox[HTML]{ffffa8}{0.0047} & \colorbox[HTML]{ffffa8}{26.12} & \colorbox[HTML]{ffffa8}{0.9152} & 0.1045 & \colorbox[HTML]{ffffa8}{0.0031} & \colorbox[HTML]{ffffa8}{26.96} & \colorbox[HTML]{ffffa8}{0.9405} & \colorbox[HTML]{ffffa8}{0.0725} \\
SuperMat             & \colorbox[HTML]{ffd7ac}{0.0024} & \colorbox[HTML]{ffd7ac}{27.66} & \colorbox[HTML]{ffd7ac}{0.9209} & \colorbox[HTML]{ffd7ac}{0.0865} & \colorbox[HTML]{ffd7ac}{0.0044} & \colorbox[HTML]{ffd7ac}{26.63} & \colorbox[HTML]{ffd7ac}{0.9171} & \colorbox[HTML]{ffaeb0}{0.0997} & \colorbox[HTML]{ffaeb0}{0.0018} & \colorbox[HTML]{ffaeb0}{28.74} & \colorbox[HTML]{ffaeb0}{0.9490} & \colorbox[HTML]{ffaeb0}{0.0597} \\
SuperMatMV           & \colorbox[HTML]{ffaeb0}{0.0022} & \colorbox[HTML]{ffaeb0}{28.04} & \colorbox[HTML]{ffaeb0}{0.9241} & \colorbox[HTML]{ffaeb0}{0.0832} & \colorbox[HTML]{ffaeb0}{0.0039} & \colorbox[HTML]{ffaeb0}{26.75} & \colorbox[HTML]{ffaeb0}{0.9183} & \colorbox[HTML]{ffffa8}{0.1026} & \colorbox[HTML]{ffd7ac}{0.0023} & \colorbox[HTML]{ffd7ac}{27.89} & \colorbox[HTML]{ffd7ac}{0.9443} & \colorbox[HTML]{ffd7ac}{0.0658} \\
\bottomrule
\end{tabular}
\vspace{-7pt}
\caption{Quantitative comparison on albedo.
We highlight the \colorbox[HTML]{ffaeb0}{best}, \colorbox[HTML]{ffd7ac}{second-best}, and \colorbox[HTML]{ffffa8}{third-best} results for each metric.}
\label{tab:2}
\vspace{-7pt}
\end{table*}

\begin{table*}[ht]
\renewcommand{\tabcolsep}{2pt}
\small
\centering
\begin{tabular}{lcccccccccccccccc}
\toprule
\multicolumn{1}{c}{} & \multicolumn{4}{c}{Objaverse}      & \multicolumn{4}{c}{BlenderVault}   & \multicolumn{4}{c}{DTC}\\ \cline{2-13} 
\multicolumn{1}{c}{} & MSE↓   & PSNR↑   & SSIM↑  & LPIPS↓ & MSE↓   & PSNR↑   & SSIM↑  & LPIPS↓ & MSE↓   & PSNR↑   & SSIM↑  & LPIPS↓ \\ \hline
IIR                  & 0.0462 & 16.87 & 0.8399 & 0.2230 & 0.0472 & 17.11 & 0.7688 & 0.2900 & 0.0324 & 19.86 & 0.6771 & 0.3604 \\
IID                  & 0.0354 & 17.31 & 0.8307 & 0.2108 & 0.0431 & 17.43 & 0.7742 & 0.2866 & 0.0347 & 17.41 & 0.6797 & 0.3895 \\
RGB$\rightarrow$X    & 0.0350 & 16.81 & 0.8296 & 0.2090 & 0.0549 & 15.36 & 0.7634 & 0.3374 & 0.0682 & 13.90 & 0.6738 & 0.4617 \\
StableMaterial       & 0.0493 & 16.83 & 0.8398 & 0.2164 & 0.0515 & 20.79 & 0.8156 & 0.2379 & 0.0567 & 23.26 & 0.7938 & 0.2833 \\
StableMaterialMV     & 0.0452 & 16.96 & 0.8411 & 0.2114 & 0.0511 & 18.74 & 0.8150 & 0.2590 & 0.0777 & 16.15 & 0.7096 & 0.3783 \\ \hline
SuperMat w/o e2e     & 0.0373 & 19.40 & \colorbox[HTML]{ffffa8}{0.8672} & 0.2021 & 0.0470 & 20.91 & \colorbox[HTML]{ffaeb0}{0.8455} & 0.2764 & 0.0496 & 22.05 & \colorbox[HTML]{ffffa8}{0.8204} & 0.3694 \\
SuperMat w/o re-render   & \colorbox[HTML]{ffffa8}{0.0135} & \colorbox[HTML]{ffffa8}{22.87} & 0.8669 & \colorbox[HTML]{ffffa8}{0.1760} & \colorbox[HTML]{ffaeb0}{0.0311} & \colorbox[HTML]{ffffa8}{22.52} & 0.8040 & \colorbox[HTML]{ffffa8}{0.2343} & \colorbox[HTML]{ffffa8}{0.0087} & \colorbox[HTML]{ffffa8}{28.23} & 0.8067 & \colorbox[HTML]{ffffa8}{0.2687} \\
SuperMat             & \colorbox[HTML]{ffd7ac}{0.0109} & \colorbox[HTML]{ffd7ac}{23.78} & \colorbox[HTML]{ffd7ac}{0.8785} & \colorbox[HTML]{ffd7ac}{0.1695} & \colorbox[HTML]{ffffa8}{0.0344} & \colorbox[HTML]{ffd7ac}{23.02} & \colorbox[HTML]{ffffa8}{0.8335} & \colorbox[HTML]{ffd7ac}{0.2328} & \colorbox[HTML]{ffd7ac}{0.0068} & \colorbox[HTML]{ffd7ac}{29.65} & \colorbox[HTML]{ffaeb0}{0.8936} & \colorbox[HTML]{ffaeb0}{0.2641} \\
SuperMatMV           & \colorbox[HTML]{ffaeb0}{0.0069} & \colorbox[HTML]{ffaeb0}{25.38} & \colorbox[HTML]{ffaeb0}{0.8919} & \colorbox[HTML]{ffaeb0}{0.1619} & \colorbox[HTML]{ffd7ac}{0.0324} & \colorbox[HTML]{ffaeb0}{23.17} & \colorbox[HTML]{ffd7ac}{0.8437} & \colorbox[HTML]{ffaeb0}{0.2311} & \colorbox[HTML]{ffaeb0}{0.0059} & \colorbox[HTML]{ffaeb0}{29.78} & \colorbox[HTML]{ffd7ac}{0.8892} & \colorbox[HTML]{ffd7ac}{0.2656} \\
\bottomrule
\end{tabular}
\vspace{-7pt}
\caption{Quantitative comparison on metallic.
We highlight the \colorbox[HTML]{ffaeb0}{best}, \colorbox[HTML]{ffd7ac}{second-best}, and \colorbox[HTML]{ffffa8}{third-best} results for each metric.}
\label{tab:3}
\vspace{-7pt}
\end{table*}

\begin{table*}[ht]
\renewcommand{\tabcolsep}{2pt}
\small
\centering
\begin{tabular}{lcccccccccccccccc}
\toprule
\multicolumn{1}{c}{} & \multicolumn{4}{c}{Objaverse}      & \multicolumn{4}{c}{BlenderVault}   & \multicolumn{4}{c}{DTC}\\ \cline{2-13} 
\multicolumn{1}{c}{} & MSE↓   & PSNR↑   & SSIM↑  & LPIPS↓ & MSE↓   & PSNR↑   & SSIM↑  & LPIPS↓ & MSE↓   & PSNR↑   & SSIM↑  & LPIPS↓ \\ \hline
IIR                  & 0.0257 & 20.67 & 0.8902 & 0.1242 & 0.0438 & 18.58 & 0.8503 & 0.1754 & 0.0320 & 19.93 & 0.8512 & 0.1793 \\
IID                  & 0.0204 & 21.32 & 0.8863 & 0.1266 & 0.0295 & 19.97 & 0.8718 & 0.1586 & 0.0168 & 21.51 & 0.8695 & 0.1576 \\
RGB$\rightarrow$X    & 0.0159 & 20.93 & 0.8705 & 0.1177 & 0.0257 & 19.78 & 0.8486 & 0.1780 & 0.0179 & 20.48 & 0.8229 & 0.1817 \\
StableMaterial       & \colorbox[HTML]{ffffa8}{0.0131} & 21.97 & 0.9065 & \colorbox[HTML]{ffaeb0}{0.1064} & 0.0251 & 20.94 & \colorbox[HTML]{ffffa8}{0.8962} & \colorbox[HTML]{ffffa8}{0.1312} & 0.0256 & 20.13 & 0.8612 & 0.1604 \\
StableMaterialMV     & 0.0142 & 21.26 & 0.8949 & \colorbox[HTML]{ffffa8}{0.1081} & 0.0253 & 20.49 & 0.8915 & \colorbox[HTML]{ffaeb0}{0.1284} & 0.0233 & 20.83 & 0.8651 & \colorbox[HTML]{ffffa8}{0.1493} \\ \hline
SuperMat w/o e2e     & 0.0170 & 21.36 & 0.8832 & 0.1221 & 0.0278 & 20.48 & 0.8679 & 0.1633 & 0.0246 & 20.60 & 0.8352 & 0.1907 \\
SuperMat w/o re-render   & \colorbox[HTML]{ffd7ac}{0.0081} & \colorbox[HTML]{ffffa8}{24.15} & \colorbox[HTML]{ffffa8}{0.9145} & 0.1156 & \colorbox[HTML]{ffffa8}{0.0211} & \colorbox[HTML]{ffffa8}{21.73} & 0.8942 & 0.1429 & \colorbox[HTML]{ffffa8}{0.0086} & \colorbox[HTML]{ffffa8}{24.69} & \colorbox[HTML]{ffffa8}{0.8961} & 0.1502 \\
SuperMat             & \colorbox[HTML]{ffaeb0}{0.0074} & \colorbox[HTML]{ffd7ac}{24.59} & \colorbox[HTML]{ffd7ac}{0.9154} & 0.1114 & \colorbox[HTML]{ffd7ac}{0.0201} & \colorbox[HTML]{ffd7ac}{22.39} & \colorbox[HTML]{ffd7ac}{0.8972} & 0.1386 & \colorbox[HTML]{ffd7ac}{0.0053} & \colorbox[HTML]{ffd7ac}{25.78} & \colorbox[HTML]{ffd7ac}{0.9046} & \colorbox[HTML]{ffd7ac}{0.1342} \\
SuperMatMV           & \colorbox[HTML]{ffaeb0}{0.0074} & \colorbox[HTML]{ffaeb0}{24.96} & \colorbox[HTML]{ffaeb0}{0.9165} & \colorbox[HTML]{ffd7ac}{0.1070} & \colorbox[HTML]{ffaeb0}{0.0145} & \colorbox[HTML]{ffaeb0}{23.22} & \colorbox[HTML]{ffaeb0}{0.8998} & \colorbox[HTML]{ffd7ac}{0.1297} & \colorbox[HTML]{ffaeb0}{0.0049} & \colorbox[HTML]{ffaeb0}{26.34} & \colorbox[HTML]{ffaeb0}{0.9064} & \colorbox[HTML]{ffaeb0}{0.1257} \\
\bottomrule
\end{tabular}
\vspace{-7pt}
\caption{Quantitative comparison on roughness.
We highlight the \colorbox[HTML]{ffaeb0}{best}, \colorbox[HTML]{ffd7ac}{second-best}, and \colorbox[HTML]{ffffa8}{third-best} results for each metric.}
\label{tab:4}
\vspace{-7pt}
\end{table*}

\begin{table*}[ht]
\renewcommand{\tabcolsep}{2pt}
\small
\centering
\begin{tabular}{lcccccccccccccccc}
\toprule
\multicolumn{1}{c}{} & \multicolumn{4}{c}{Objaverse}      & \multicolumn{4}{c}{BlenderVault}   & \multicolumn{4}{c}{DTC}\\ \cline{2-13} 
\multicolumn{1}{c}{} & MSE↓   & PSNR↑   & SSIM↑  & LPIPS↓ & MSE↓   & PSNR↑   & SSIM↑  & LPIPS↓ & MSE↓   & PSNR↑   & SSIM↑  & LPIPS↓ \\ \hline
IIR                  & 0.0132 & 23.19 & 0.8991 & 0.1060 & 0.0290 & 20.96 & 0.8765 & 0.1441 & 0.0245 & 18.80 & 0.8897 & 0.0890 \\
IID                  & 0.0105 & 23.83 & 0.9057 & 0.0971 & 0.0160 & 22.31 & 0.9009 & 0.1142 & 0.0149 & 21.21 & 0.9066 & 0.1137 \\
RGB$\rightarrow$X    & 0.0107 & 22.86 & 0.8974 & 0.0864 & 0.0177 & 21.43 & 0.8828 & 0.1205 & 0.0207 & 20.23 & 0.8780 & 0.1327 \\
StableMaterial       & 0.0107 & 24.08 & 0.9108 & 0.0851 & 0.0203 & 22.22 & 0.9022 & 0.1013 & 0.0183 & 21.38 & 0.8999 & 0.1074 \\
StableMaterialMV     & 0.0091 & 24.02 & 0.9114 & 0.0816 & 0.0154 & 22.69 & 0.9019 & 0.1018 & 0.0134 & 22.70 & 0.9137 & 0.0966 \\ \hline
SuperMat w/o e2e     & 0.0079 & 25.81 & 0.9192 & \colorbox[HTML]{ffffa8}{0.0776} & 0.0171 & 23.15 & 0.9031 & 0.1059 & 0.0158 & 22.75 & 0.9032 & 0.1057 \\
SuperMat w/o re-render   & \colorbox[HTML]{ffffa8}{0.0042} & \colorbox[HTML]{ffffa8}{27.25} & \colorbox[HTML]{ffffa8}{0.9374} & \colorbox[HTML]{ffd7ac}{0.0650} & \colorbox[HTML]{ffd7ac}{0.0109} & \colorbox[HTML]{ffffa8}{24.94} & \colorbox[HTML]{ffffa8}{0.9270} & \colorbox[HTML]{ffffa8}{0.0855} & \colorbox[HTML]{ffffa8}{0.0050} & \colorbox[HTML]{ffffa8}{27.05} & \colorbox[HTML]{ffffa8}{0.9490} & \colorbox[HTML]{ffffa8}{0.0600} \\
SuperMat             & \colorbox[HTML]{ffd7ac}{0.0041} & \colorbox[HTML]{ffd7ac}{28.01} & \colorbox[HTML]{ffd7ac}{0.9406} & \colorbox[HTML]{ffaeb0}{0.0566} & \colorbox[HTML]{ffaeb0}{0.0101} & \colorbox[HTML]{ffaeb0}{25.50} & \colorbox[HTML]{ffaeb0}{0.9300} & \colorbox[HTML]{ffaeb0}{0.0811} & \colorbox[HTML]{ffaeb0}{0.0031} & \colorbox[HTML]{ffaeb0}{29.47} & \colorbox[HTML]{ffaeb0}{0.9590} & \colorbox[HTML]{ffaeb0}{0.0460} \\
SuperMatMV           & \colorbox[HTML]{ffaeb0}{0.0032} & \colorbox[HTML]{ffaeb0}{28.51} & \colorbox[HTML]{ffaeb0}{0.9437} & \colorbox[HTML]{ffaeb0}{0.0566} & \colorbox[HTML]{ffffa8}{0.0118} & \colorbox[HTML]{ffd7ac}{25.41} & \colorbox[HTML]{ffd7ac}{0.9289} & \colorbox[HTML]{ffd7ac}{0.0815} & \colorbox[HTML]{ffd7ac}{0.0041} & \colorbox[HTML]{ffd7ac}{29.01} & \colorbox[HTML]{ffd7ac}{0.9567} & \colorbox[HTML]{ffd7ac}{0.0502} \\
\bottomrule
\end{tabular}
\vspace{-7pt}
\caption{Quantitative comparison on relighting.
We highlight the \colorbox[HTML]{ffaeb0}{best}, \colorbox[HTML]{ffd7ac}{second-best}, and \colorbox[HTML]{ffffa8}{third-best} results for each metric.}
\label{tab:5}
\end{table*}

%% file: main.bbl
\begin{thebibliography}{67}
\providecommand{\natexlab}[1]{#1}
\providecommand{\url}[1]{\texttt{#1}}
\expandafter\ifx\csname urlstyle\endcsname\relax
  \providecommand{\doi}[1]{doi: #1}\else
  \providecommand{\doi}{doi: \begingroup \urlstyle{rm}\Url}\fi

\bibitem[Bensadoun et~al.(2024)Bensadoun, Kleiman, Azuri, Harosh, Vedaldi, Neverova, and Gafni]{bensadoun2024meta}
Raphael Bensadoun, Yanir Kleiman, Idan Azuri, Omri Harosh, Andrea Vedaldi, Natalia Neverova, and Oran Gafni.
\newblock Meta 3d texturegen: Fast and consistent texture generation for 3d objects.
\newblock \emph{arXiv preprint arXiv:2407.02430}, 2024.

\bibitem[Boss et~al.(2021)Boss, Jampani, Braun, Liu, Barron, and Lensch]{boss2021neural}
Mark Boss, Varun Jampani, Raphael Braun, Ce Liu, Jonathan Barron, and Hendrik Lensch.
\newblock Neural-pil: Neural pre-integrated lighting for reflectance decomposition.
\newblock \emph{NeurIPS}, 2021.

\bibitem[Boss et~al.(2024)Boss, Huang, Vasishta, and Jampani]{boss2024sf3d}
Mark Boss, Zixuan Huang, Aaryaman Vasishta, and Varun Jampani.
\newblock Sf3d: Stable fast 3d mesh reconstruction with uv-unwrapping and illumination disentanglement.
\newblock \emph{arXiv preprint arXiv:2408.00653}, 2024.

\bibitem[Chen et~al.(2023{\natexlab{a}})Chen, Siddiqui, Lee, Tulyakov, and Nie{\ss}ner]{chen2023text2tex}
Dave~Zhenyu Chen, Yawar Siddiqui, Hsin-Ying Lee, Sergey Tulyakov, and Matthias Nie{\ss}ner.
\newblock Text2tex: Text-driven texture synthesis via diffusion models.
\newblock In \emph{ICCV}, 2023{\natexlab{a}}.

\bibitem[Chen et~al.(2023{\natexlab{b}})Chen, Chen, Jiao, and Jia]{chen2023fantasia3d}
Rui Chen, Yongwei Chen, Ningxin Jiao, and Kui Jia.
\newblock Fantasia3d: Disentangling geometry and appearance for high-quality text-to-3d content creation.
\newblock In \emph{ICCV}, 2023{\natexlab{b}}.

\bibitem[Chen et~al.(2024)Chen, Peng, Yang, Liu, Pan, Lv, and Zhou]{chen2024intrinsicanything}
Xi Chen, Sida Peng, Dongchen Yang, Yuan Liu, Bowen Pan, Chengfei Lv, and Xiaowei Zhou.
\newblock Intrinsicanything: Learning diffusion priors for inverse rendering under unknown illumination.
\newblock \emph{arXiv preprint arXiv:2404.11593}, 2024.

\bibitem[Cheng et~al.(2023)Cheng, Li, and Li]{cheng2023wildlight}
Ziang Cheng, Junxuan Li, and Hongdong Li.
\newblock Wildlight: In-the-wild inverse rendering with a flashlight.
\newblock In \emph{CVPR}, 2023.

\bibitem[Cook and Torrance(1982)]{cook1982reflectance}
Robert~L Cook and Kenneth~E. Torrance.
\newblock A reflectance model for computer graphics.
\newblock \emph{TOG}, 1982.

\bibitem[Deitke et~al.(2023)Deitke, Schwenk, Salvador, Weihs, Michel, VanderBilt, Schmidt, Ehsani, Kembhavi, and Farhadi]{deitke2023objaverse}
Matt Deitke, Dustin Schwenk, Jordi Salvador, Luca Weihs, Oscar Michel, Eli VanderBilt, Ludwig Schmidt, Kiana Ehsani, Aniruddha Kembhavi, and Ali Farhadi.
\newblock Objaverse: A universe of annotated 3d objects.
\newblock In \emph{CVPR}, 2023.

\bibitem[Deng et~al.(2024)Deng, Omernick, Weiss, Ramanan, Zhu, Zhou, and Agrawala]{deng2024flashtex}
Kangle Deng, Timothy Omernick, Alexander Weiss, Deva Ramanan, Jun-Yan Zhu, Tinghui Zhou, and Maneesh Agrawala.
\newblock Flashtex: Fast relightable mesh texturing with lightcontrolnet.
\newblock In \emph{ECCV}, 2024.

\bibitem[Deschaintre et~al.(2018)Deschaintre, Aittala, Durand, Drettakis, and Bousseau]{deschaintre2018single}
Valentin Deschaintre, Miika Aittala, Fredo Durand, George Drettakis, and Adrien Bousseau.
\newblock Single-image svbrdf capture with a rendering-aware deep network.
\newblock \emph{TOG}, 2018.

\bibitem[Duan et~al.(2023)Duan, Guo, and Zhu]{duan2023diffusiondepth}
Yiqun Duan, Xianda Guo, and Zheng Zhu.
\newblock Diffusiondepth: Diffusion denoising approach for monocular depth estimation.
\newblock \emph{arXiv preprint arXiv:2303.05021}, 2023.

\bibitem[Feng et~al.(2024)Feng, Yu, Bi, Shang, Gao, Wu, Zhou, Jiang, and Yang]{feng2024arm}
Xiang Feng, Chang Yu, Zoubin Bi, Yintong Shang, Feng Gao, Hongzhi Wu, Kun Zhou, Chenfanfu Jiang, and Yin Yang.
\newblock Arm: Appearance reconstruction model for relightable 3d generation.
\newblock \emph{arXiv preprint arXiv:2411.10825}, 2024.

\bibitem[Fort(2024)]{dtc}
James Fort.
\newblock Introducing the digital twin catalog from reality labs research, 2024.
\newblock \url{https://ai.meta.com/blog/digital-twin-catalog-3d-reconstruction-shopify-reality-labs-research/}, Last accessed on 2025-03-07.

\bibitem[Fu et~al.(2025)Fu, Yin, Hu, Wang, Ma, Tan, Shen, Lin, and Long]{fu2025geowizard}
Xiao Fu, Wei Yin, Mu Hu, Kaixuan Wang, Yuexin Ma, Ping Tan, Shaojie Shen, Dahua Lin, and Xiaoxiao Long.
\newblock Geowizard: Unleashing the diffusion priors for 3d geometry estimation from a single image.
\newblock In \emph{ECCV}, 2025.

\bibitem[Gao et~al.(2023)Gao, Gu, Lin, Zhu, Cao, Zhang, and Yao]{gao2023relightable}
Jian Gao, Chun Gu, Youtian Lin, Hao Zhu, Xun Cao, Li Zhang, and Yao Yao.
\newblock Relightable 3d gaussian: Real-time point cloud relighting with brdf decomposition and ray tracing.
\newblock \emph{arXiv preprint arXiv:2311.16043}, 2023.

\bibitem[Gao et~al.(2021)Gao, Wu, Yuan, Lin, Lai, and Zhang]{gao2021tm}
Lin Gao, Tong Wu, Yu-Jie Yuan, Ming-Xian Lin, Yu-Kun Lai, and Hao Zhang.
\newblock Tm-net: Deep generative networks for textured meshes.
\newblock \emph{TOG}, 2021.

\bibitem[Garcia et~al.(2024)Garcia, Zeid, Schmidt, de~Geus, Hermans, and Leibe]{garcia2024fine}
Gonzalo~Martin Garcia, Karim~Abou Zeid, Christian Schmidt, Daan de Geus, Alexander Hermans, and Bastian Leibe.
\newblock Fine-tuning image-conditional diffusion models is easier than you think.
\newblock \emph{arXiv preprint arXiv:2409.11355}, 2024.

\bibitem[Gui et~al.(2024)Gui, Fischer, Prestel, Ma, Kotovenko, Grebenkova, Baumann, Hu, and Ommer]{gui2024depthfm}
Ming Gui, Johannes~S Fischer, Ulrich Prestel, Pingchuan Ma, Dmytro Kotovenko, Olga Grebenkova, Stefan~Andreas Baumann, Vincent~Tao Hu, and Bj{\"o}rn Ommer.
\newblock Depthfm: Fast monocular depth estimation with flow matching.
\newblock \emph{arXiv preprint arXiv:2403.13788}, 2024.

\bibitem[Guo et~al.(2022)Guo, Kang, Bao, He, and Zhang]{guo2022nerfren}
Yuan-Chen Guo, Di Kang, Linchao Bao, Yu He, and Song-Hai Zhang.
\newblock Nerfren: Neural radiance fields with reflections.
\newblock In \emph{CVPR}, 2022.

\bibitem[Hasselgren et~al.(2022)Hasselgren, Hofmann, and Munkberg]{hasselgren2022shape}
Jon Hasselgren, Nikolai Hofmann, and Jacob Munkberg.
\newblock Shape, light, and material decomposition from images using monte carlo rendering and denoising.
\newblock \emph{NeurIPS}, 2022.

\bibitem[He et~al.(2024)He, Li, Yin, Liang, Li, Zhou, Zhang, Liu, and Chen]{he2024lotus}
Jing He, Haodong Li, Wei Yin, Yixun Liang, Leheng Li, Kaiqiang Zhou, Hongbo Zhang, Bingbing Liu, and Ying-Cong Chen.
\newblock Lotus: Diffusion-based visual foundation model for high-quality dense prediction.
\newblock \emph{arXiv preprint arXiv:2409.18124}, 2024.

\bibitem[Jin et~al.(2023)Jin, Liu, Xu, Zhang, Han, Bi, Zhou, Xu, and Su]{jin2023tensoir}
Haian Jin, Isabella Liu, Peijia Xu, Xiaoshuai Zhang, Songfang Han, Sai Bi, Xiaowei Zhou, Zexiang Xu, and Hao Su.
\newblock Tensoir: Tensorial inverse rendering.
\newblock In \emph{CVPR}, 2023.

\bibitem[Johnson et~al.(2016)Johnson, Alahi, and Fei-Fei]{johnson2016perceptual}
Justin Johnson, Alexandre Alahi, and Li Fei-Fei.
\newblock Perceptual losses for real-time style transfer and super-resolution.
\newblock In \emph{ECCV}, 2016.

\bibitem[Ke et~al.(2024)Ke, Obukhov, Huang, Metzger, Daudt, and Schindler]{ke2024repurposing}
Bingxin Ke, Anton Obukhov, Shengyu Huang, Nando Metzger, Rodrigo~Caye Daudt, and Konrad Schindler.
\newblock Repurposing diffusion-based image generators for monocular depth estimation.
\newblock In \emph{CVPR}, 2024.

\bibitem[Kocsis et~al.(2024)Kocsis, Sitzmann, and Nie{\ss}ner]{kocsis2024intrinsic}
Peter Kocsis, Vincent Sitzmann, and Matthias Nie{\ss}ner.
\newblock Intrinsic image diffusion for indoor single-view material estimation.
\newblock In \emph{CVPR}, 2024.

\bibitem[Kuang et~al.(2023)Kuang, Zhang, Yu, Agarwala, Wu, Wu, et~al.]{kuang2023stanfordorb}
Zhengfei Kuang, Yunzhi Zhang, Hong-Xing Yu, Samir Agarwala, Elliott Wu, Jiajun Wu, et~al.
\newblock Stanford-orb: a real-world 3d object inverse rendering benchmark.
\newblock In \emph{NeurIPS}, 2023.

\bibitem[Liang et~al.(2024)Liang, Zhang, Feng, Shan, and Jia]{liang2024gs}
Zhihao Liang, Qi Zhang, Ying Feng, Ying Shan, and Kui Jia.
\newblock Gs-ir: 3d gaussian splatting for inverse rendering.
\newblock In \emph{CVPR}, 2024.

\bibitem[Lin et~al.(2024)Lin, Liu, Li, and Yang]{lin2024common}
Shanchuan Lin, Bingchen Liu, Jiashi Li, and Xiao Yang.
\newblock Common diffusion noise schedules and sample steps are flawed.
\newblock In \emph{WACV}, 2024.

\bibitem[Litman et~al.(2024)Litman, Patashnik, Deng, Agrawal, Zawar, De~la Torre, and Tulsiani]{litman2024materialfusion}
Yehonathan Litman, Or Patashnik, Kangle Deng, Aviral Agrawal, Rushikesh Zawar, Fernando De~la Torre, and Shubham Tulsiani.
\newblock Materialfusion: Enhancing inverse rendering with material diffusion priors.
\newblock \emph{arXiv preprint arXiv:2409.15273}, 2024.

\bibitem[Liu et~al.(2023{\natexlab{a}})Liu, Ren, Siarohin, Skorokhodov, Li, Lin, Liu, Liu, and Tulyakov]{liu2023hyperhuman}
Xian Liu, Jian Ren, Aliaksandr Siarohin, Ivan Skorokhodov, Yanyu Li, Dahua Lin, Xihui Liu, Ziwei Liu, and Sergey Tulyakov.
\newblock Hyperhuman: Hyper-realistic human generation with latent structural diffusion.
\newblock \emph{arXiv preprint arXiv:2310.08579}, 2023{\natexlab{a}}.

\bibitem[Liu et~al.(2020)Liu, Li, You, and Lu]{liu2020unsupervised}
Yunfei Liu, Yu Li, Shaodi You, and Feng Lu.
\newblock Unsupervised learning for intrinsic image decomposition from a single image.
\newblock In \emph{CVPR}, 2020.

\bibitem[Liu et~al.(2023{\natexlab{b}})Liu, Xie, Liu, and Wong]{liu2023text}
Yuxin Liu, Minshan Xie, Hanyuan Liu, and Tien-Tsin Wong.
\newblock Text-guided texturing by synchronized multi-view diffusion.
\newblock \emph{arXiv preprint arXiv:2311.12891}, 2023{\natexlab{b}}.

\bibitem[Liu et~al.(2024)Liu, Li, Lin, Yu, Peng, Cao, Qi, Huang, Liang, and Ouyang]{liu2024unidream}
Zexiang Liu, Yangguang Li, Youtian Lin, Xin Yu, Sida Peng, Yan-Pei Cao, Xiaojuan Qi, Xiaoshui Huang, Ding Liang, and Wanli Ouyang.
\newblock Unidream: Unifying diffusion priors for relightable text-to-3d generation.
\newblock In \emph{ECCV}, 2024.

\bibitem[Mohammad~Khalid et~al.(2022)Mohammad~Khalid, Xie, Belilovsky, and Popa]{mohammad2022clip}
Nasir Mohammad~Khalid, Tianhao Xie, Eugene Belilovsky, and Tiberiu Popa.
\newblock Clip-mesh: Generating textured meshes from text using pretrained image-text models.
\newblock In \emph{SIGGRAPH Asia}, 2022.

\bibitem[Munkberg et~al.(2022)Munkberg, Hasselgren, Shen, Gao, Chen, Evans, M{\"u}ller, and Fidler]{munkberg2022extracting}
Jacob Munkberg, Jon Hasselgren, Tianchang Shen, Jun Gao, Wenzheng Chen, Alex Evans, Thomas M{\"u}ller, and Sanja Fidler.
\newblock Extracting triangular 3d models, materials, and lighting from images.
\newblock In \emph{CVPR}, 2022.

\bibitem[Oquab et~al.(2023)Oquab, Darcet, Moutakanni, Vo, Szafraniec, Khalidov, Fernandez, Haziza, Massa, El-Nouby, et~al.]{oquab2023dinov2}
Maxime Oquab, Timoth{\'e}e Darcet, Th{\'e}o Moutakanni, Huy Vo, Marc Szafraniec, Vasil Khalidov, Pierre Fernandez, Daniel Haziza, Francisco Massa, Alaaeldin El-Nouby, et~al.
\newblock Dinov2: Learning robust visual features without supervision.
\newblock \emph{arXiv preprint arXiv:2304.07193}, 2023.

\bibitem[Poole et~al.(2022)Poole, Jain, Barron, and Mildenhall]{poole2022dreamfusion}
Ben Poole, Ajay Jain, Jonathan~T Barron, and Ben Mildenhall.
\newblock Dreamfusion: Text-to-3d using 2d diffusion.
\newblock \emph{arXiv preprint arXiv:2209.14988}, 2022.

\bibitem[Qiu et~al.(2024)Qiu, Chen, Gu, Zuo, Xu, Wu, Yuan, Dong, Bo, and Han]{qiu2024richdreamer}
Lingteng Qiu, Guanying Chen, Xiaodong Gu, Qi Zuo, Mutian Xu, Yushuang Wu, Weihao Yuan, Zilong Dong, Liefeng Bo, and Xiaoguang Han.
\newblock Richdreamer: A generalizable normal-depth diffusion model for detail richness in text-to-3d.
\newblock In \emph{CVPR}, 2024.

\bibitem[Ranftl et~al.(2020)Ranftl, Lasinger, Hafner, Schindler, and Koltun]{ranftl2020towards}
Ren{\'e} Ranftl, Katrin Lasinger, David Hafner, Konrad Schindler, and Vladlen Koltun.
\newblock Towards robust monocular depth estimation: Mixing datasets for zero-shot cross-dataset transfer.
\newblock \emph{PAMI}, 2020.

\bibitem[Richardson et~al.(2023)Richardson, Metzer, Alaluf, Giryes, and Cohen-Or]{richardson2023texture}
Elad Richardson, Gal Metzer, Yuval Alaluf, Raja Giryes, and Daniel Cohen-Or.
\newblock Texture: Text-guided texturing of 3d shapes.
\newblock In \emph{SIGGRAPH}, 2023.

\bibitem[Rombach et~al.(2022)Rombach, Blattmann, Lorenz, Esser, and Ommer]{rombach2022high}
Robin Rombach, Andreas Blattmann, Dominik Lorenz, Patrick Esser, and Bj{\"o}rn Ommer.
\newblock High-resolution image synthesis with latent diffusion models.
\newblock In \emph{CVPR}, 2022.

\bibitem[Saxena et~al.(2023)Saxena, Kar, Norouzi, and Fleet]{saxena2023monocular}
Saurabh Saxena, Abhishek Kar, Mohammad Norouzi, and David~J Fleet.
\newblock Monocular depth estimation using diffusion models.
\newblock \emph{arXiv preprint arXiv:2302.14816}, 2023.

\bibitem[Shi et~al.(2023{\natexlab{a}})Shi, Wang, Ye, Long, Li, and Yang]{shi2023mvdream}
Yichun Shi, Peng Wang, Jianglong Ye, Mai Long, Kejie Li, and Xiao Yang.
\newblock Mvdream: Multi-view diffusion for 3d generation.
\newblock \emph{arXiv preprint arXiv:2308.16512}, 2023{\natexlab{a}}.

\bibitem[Shi et~al.(2023{\natexlab{b}})Shi, Lin, and Song]{shi2023attention}
Zeqi Shi, Xiangyu Lin, and Ying Song.
\newblock An attention-embedded gan for svbrdf recovery from a single image.
\newblock \emph{CVM}, 2023{\natexlab{b}}.

\bibitem[Vecchio et~al.(2024)Vecchio, Martin, Roullier, Kaiser, Rouffet, Deschaintre, and Boubekeur]{vecchio2024controlmat}
Giuseppe Vecchio, Rosalie Martin, Arthur Roullier, Adrien Kaiser, Romain Rouffet, Valentin Deschaintre, and Tamy Boubekeur.
\newblock Controlmat: a controlled generative approach to material capture.
\newblock \emph{TOG}, 2024.

\bibitem[Wang et~al.(2024{\natexlab{a}})Wang, Xu, Ma, Wang, and Dai]{wang2024boosting}
Yitong Wang, Xudong Xu, Li Ma, Haoran Wang, and Bo Dai.
\newblock Boosting 3d object generation through pbr materials.
\newblock In \emph{SIGGRAPH Asia}, 2024{\natexlab{a}}.

\bibitem[Wang et~al.(2024{\natexlab{b}})Wang, Lu, Wang, Bao, Li, Su, and Zhu]{wang2024prolificdreamer}
Zhengyi Wang, Cheng Lu, Yikai Wang, Fan Bao, Chongxuan Li, Hang Su, and Jun Zhu.
\newblock Prolificdreamer: High-fidelity and diverse text-to-3d generation with variational score distillation.
\newblock \emph{NeurIPS}, 2024{\natexlab{b}}.

\bibitem[Wimbauer et~al.(2022)Wimbauer, Wu, and Rupprecht]{wimbauer2022rendering}
Felix Wimbauer, Shangzhe Wu, and Christian Rupprecht.
\newblock De-rendering 3d objects in the wild.
\newblock In \emph{CVPR}, 2022.

\bibitem[Xiong et~al.(2024)Xiong, Liu, Hu, Wu, Wu, Liu, Zhao, Ding, and Lian]{xiong2024texgaussian}
Bojun Xiong, Jialun Liu, Jiakui Hu, Chenming Wu, Jinbo Wu, Xing Liu, Chen Zhao, Errui Ding, and Zhouhui Lian.
\newblock Texgaussian: Generating high-quality pbr material via octree-based 3d gaussian splatting.
\newblock \emph{arXiv preprint arXiv:2411.19654}, 2024.

\bibitem[Xu et~al.(2024)Xu, Ge, Liu, Fan, Xie, Zhao, Chen, and Shen]{xu2024diffusion}
Guangkai Xu, Yongtao Ge, Mingyu Liu, Chengxiang Fan, Kangyang Xie, Zhiyue Zhao, Hao Chen, and Chunhua Shen.
\newblock Diffusion models trained with large data are transferable visual models.
\newblock \emph{arXiv preprint arXiv:2403.06090}, 2024.

\bibitem[Xu et~al.(2023)Xu, Lyu, Pan, and Dai]{xu2023matlaber}
Xudong Xu, Zhaoyang Lyu, Xingang Pan, and Bo Dai.
\newblock Matlaber: Material-aware text-to-3d via latent brdf auto-encoder.
\newblock \emph{arXiv preprint arXiv:2308.09278}, 2023.

\bibitem[Ye et~al.(2024)Ye, Qiu, Gu, Zuo, Wu, Dong, Bo, Xiu, and Han]{ye2024stablenormal}
Chongjie Ye, Lingteng Qiu, Xiaodong Gu, Qi Zuo, Yushuang Wu, Zilong Dong, Liefeng Bo, Yuliang Xiu, and Xiaoguang Han.
\newblock Stablenormal: Reducing diffusion variance for stable and sharp normal.
\newblock \emph{arXiv preprint arXiv:2406.16864}, 2024.

\bibitem[Yi et~al.(2023)Yi, Zhu, and Xu]{yi2023weakly}
Renjiao Yi, Chenyang Zhu, and Kai Xu.
\newblock Weakly-supervised single-view image relighting.
\newblock In \emph{CVPR}, 2023.

\bibitem[Youwang et~al.(2024)Youwang, Oh, and Pons-Moll]{youwang2024paint}
Kim Youwang, Tae-Hyun Oh, and Gerard Pons-Moll.
\newblock Paint-it: Text-to-texture synthesis via deep convolutional texture map optimization and physically-based rendering.
\newblock In \emph{CVPR}, 2024.

\bibitem[Yu et~al.(2023)Yu, Dai, Li, Ma, Liu, and Qi]{yu2023texture}
Xin Yu, Peng Dai, Wenbo Li, Lan Ma, Zhengzhe Liu, and Xiaojuan Qi.
\newblock Texture generation on 3d meshes with point-uv diffusion.
\newblock In \emph{ICCV}, 2023.

\bibitem[Yu et~al.(2024)Yu, Yuan, Guo, Liu, Liu, Li, Cao, Liang, and Qi]{yu2024texgen}
Xin Yu, Ze Yuan, Yuan-Chen Guo, Ying-Tian Liu, Jianhui Liu, Yangguang Li, Yan-Pei Cao, Ding Liang, and Xiaojuan Qi.
\newblock Texgen: a generative diffusion model for mesh textures.
\newblock \emph{TOG}, 2024.

\bibitem[Zeng et~al.(2024{\natexlab{a}})Zeng, Chen, Qi, Liu, Zhao, Wang, Fu, Liu, and Yu]{zeng2024paint3d}
Xianfang Zeng, Xin Chen, Zhongqi Qi, Wen Liu, Zibo Zhao, Zhibin Wang, Bin Fu, Yong Liu, and Gang Yu.
\newblock Paint3d: Paint anything 3d with lighting-less texture diffusion models.
\newblock In \emph{CVPR}, 2024{\natexlab{a}}.

\bibitem[Zeng et~al.(2024{\natexlab{b}})Zeng, Deschaintre, Georgiev, Hold-Geoffroy, Hu, Luan, Yan, and Ha{\v{s}}an]{zeng2024rgb}
Zheng Zeng, Valentin Deschaintre, Iliyan Georgiev, Yannick Hold-Geoffroy, Yiwei Hu, Fujun Luan, Ling-Qi Yan, and Milo{\v{s}} Ha{\v{s}}an.
\newblock Rgb $\leftrightarrow$ x: Image decomposition and synthesis using material-and lighting-aware diffusion models.
\newblock In \emph{SIGGRAPH}, 2024{\natexlab{b}}.

\bibitem[Zhang et~al.(2022{\natexlab{a}})Zhang, Luan, Li, and Snavely]{zhang2022iron}
Kai Zhang, Fujun Luan, Zhengqi Li, and Noah Snavely.
\newblock Iron: Inverse rendering by optimizing neural sdfs and materials from photometric images.
\newblock In \emph{CVPR}, 2022{\natexlab{a}}.

\bibitem[Zhang et~al.(2024{\natexlab{a}})Zhang, Wang, Zhang, Qiu, Pang, Jiang, Yang, Xu, and Yu]{zhang2024clay}
Longwen Zhang, Ziyu Wang, Qixuan Zhang, Qiwei Qiu, Anqi Pang, Haoran Jiang, Wei Yang, Lan Xu, and Jingyi Yu.
\newblock Clay: A controllable large-scale generative model for creating high-quality 3d assets.
\newblock \emph{TOG}, 2024{\natexlab{a}}.

\bibitem[Zhang et~al.(2024{\natexlab{b}})Zhang, Thamizharasan, and Tompkin]{zhang2024learning}
Qian Zhang, Vikas Thamizharasan, and James Tompkin.
\newblock Learning physically based material and lighting decompositions for face editing.
\newblock \emph{CVM}, 2024{\natexlab{b}}.

\bibitem[Zhang et~al.(2021)Zhang, Srinivasan, Deng, Debevec, Freeman, and Barron]{zhang2021nerfactor}
Xiuming Zhang, Pratul~P Srinivasan, Boyang Deng, Paul Debevec, William~T Freeman, and Jonathan~T Barron.
\newblock Nerfactor: Neural factorization of shape and reflectance under an unknown illumination.
\newblock \emph{TOG}, 2021.

\bibitem[Zhang et~al.(2022{\natexlab{b}})Zhang, Sun, He, Fu, Jia, and Zhou]{zhang2022modeling}
Yuanqing Zhang, Jiaming Sun, Xingyi He, Huan Fu, Rongfei Jia, and Xiaowei Zhou.
\newblock Modeling indirect illumination for inverse rendering.
\newblock In \emph{CVPR}, 2022{\natexlab{b}}.

\bibitem[Zhang et~al.(2024{\natexlab{c}})Zhang, Liu, Xie, Yang, Liu, Yang, Zhang, Kou, Lin, Wang, et~al.]{zhang2024dreammat}
Yuqing Zhang, Yuan Liu, Zhiyu Xie, Lei Yang, Zhongyuan Liu, Mengzhou Yang, Runze Zhang, Qilong Kou, Cheng Lin, Wenping Wang, et~al.
\newblock Dreammat: High-quality pbr material generation with geometry-and light-aware diffusion models.
\newblock \emph{TOG}, 2024{\natexlab{c}}.

\bibitem[Zhao et~al.(2023)Zhao, Rao, Liu, Liu, Zhou, and Lu]{zhao2023unleashing}
Wenliang Zhao, Yongming Rao, Zuyan Liu, Benlin Liu, Jie Zhou, and Jiwen Lu.
\newblock Unleashing text-to-image diffusion models for visual perception.
\newblock In \emph{ICCV}, 2023.

\bibitem[Zhu et~al.(2022)Zhu, Luan, Huo, Lin, Zhong, Xi, Wang, Bao, Zheng, and Tang]{zhu2022learning}
Jingsen Zhu, Fujun Luan, Yuchi Huo, Zihao Lin, Zhihua Zhong, Dianbing Xi, Rui Wang, Hujun Bao, Jiaxiang Zheng, and Rui Tang.
\newblock Learning-based inverse rendering of complex indoor scenes with differentiable monte carlo raytracing.
\newblock In \emph{SIGGRAPH Asia}, 2022.

\end{thebibliography}
